\documentclass[preprint,12pt]{elsarticle}

\usepackage{amssymb}
\usepackage{orcidlink}
\usepackage{rotating}
\usepackage{lineno} 
\usepackage{siunitx}
\usepackage{multirow}
\usepackage{float}
\usepackage{booktabs}
\usepackage{threeparttable}
\usepackage{pdflscape}
\usepackage{adjustbox}
\usepackage{pifont}
\usepackage[table]{xcolor}  
\definecolor{TableStripe}{gray}{0.95}
\definecolor{TableSection}{gray}{0.86}
\definecolor{TableHeader}{gray}{0.90}
\usepackage{xurl}

\usepackage{amsmath}

\journal{Journal of Cardiovascular Magnetic Resonance}

\begin{document}

\begin{frontmatter}



\title{Segmentation of the aorta in 4D flow MRI using 4D convolutional kernels and learning from sparse annotations} 


\author[ICM,FSCPC,Charite,DZHK]{Hinrich Rahlfs\orcidlink{0009-0005-1870-9290}}
\author[UofC]{Julio Garcia}
\author[ICM,Charite]{Chiara Manini\orcidlink{0000-0001-5357-3396}}
\author[ICM,Charite,DZHK,MEVIS]{Markus Hüllebrand\orcidlink{0000-0003-4948-0917}}
\author[PTB]{Sebastian Schmitter\orcidlink{0000-0003-4410-6790}}
\author[TB]{Sarah Nordmeyer}
\author[ICM,Charite]{Titus Kühne\orcidlink{0009-0002-1849-4013}}
\author[DHZM]{Heiko Stern\orcidlink{0000-0002-9497-1638}}
\author[DHZM]{Christian Meierhofer\orcidlink{0000-0002-1367-3410}}
\author[FB]{Andreas Harloff\orcidlink{0000-0002-3252-7910}}
\author[Charite,DZHK,DHZC]{Sebastian Kelle\orcidlink{0000-0001-8105-6599}}
\author[Eppendorf]{Alexander Lenz\orcidlink{0000-0003-1916-1317}}
\author[Eppendorf]{Peter Bannas\orcidlink{0000-0002-7102-534X}}
\author[DZHK,DHZC,ECRC,HELIOS]{Jeanette Schulz-Menger\orcidlink{0000-0003-3100-1092}}
\author[DZHK,DHZC,ECRC,HELIOS,ANAE]{Ralf F Trauzeddel\orcidlink{0000-0003-3827-8895}}
\author[ICM,Charite,DZHK,MEVIS]{Anja Hennemuth\orcidlink{0000-0002-0737-7375}}

\affiliation[ICM]{organization={Deutsches Herzzentrum der Charité, Institute of Computer-assisted Cardiovascular Medicine},
            city={Berlin},
            country={Germany}}

\affiliation[FSCPC]{organization={Friede Springer Cardiovascular Prevention Center at Charité},
            city={Berlin},
            country={Germany}}
\affiliation[Charite]{organization={Charité – Universitätsmedizin Berlin, corporate member of Freie Universität Berlin and Humboldt-Universität zu Berlin},
            city={Berlin},
            country={Germany}}
\affiliation[DZHK]{organization={DZHK (German Centre for Cardiovascular Research)},
city={Partner Site Berlin},
country={Germany}}
\affiliation[UofC]{organization={Stephenson Cardiac Imaging Centre, University of Calgary},
            city={Calgary},
            country={Canada}}
\affiliation[MEVIS]{organization={Fraunhofer MEVIS},
city={Bremen},
country={Germany}}
\affiliation[PTB]{organization={Physikalisch-Technische Bundesanstalt (PTB)},
city={Berlin},
country={Germany}}
\affiliation[TB]{organization={University Hospital Tuebingen - Diagnostic and Interventional Radiology},city={Tübingen},
country={Germany}}
\affiliation[DHZM]{organization={Congenital Heart Disease and Pediatric Cardiology, German Heart Center Munich},city={Munich},
country={Germany}}
\affiliation[FB]{organization={Department of Neurology and Neurophysiology, Faculty of Medicine, Medical Center—University of Freiburg},
city={Freiburg},
country={Germany}}
\affiliation[DHZC]{organization={Deutsches Herzzentrum der Charité, Department of Cardiology, Angiology and Intensive Care Medicine},
city={Berlin},
country={Germany}}
\affiliation[Eppendorf]{organization={Department of Diagnostic and Interventional Radiology and Nuclear Medicine, University Medical Center Hamburg-Eppendorf},
city={Hamburg},
country={Germany}}
\affiliation[ECRC]{
organization={Working Group on Cardiovascular Magnetic Resonance, Experimental and Clinical Research Center, a joint cooperation between the Charite Medical Faculty and the Max-Delbrueck Center for Molecular Medicine and HELIOS Hospital Berlin-Buch, Department of Cardiology and Nephrology, Medical University Berlin, Charite Campus Buch},
city={Berlin},
country={Germany}
}
\affiliation[HELIOS]{
organization={HELIOS Hospital Berlin-Buch, Department of Cardiology and Nephrology},
city={Berlin},
country={Germany}
}
\affiliation[ANAE]{
organization={Department of Anesthesiology and Intensive Care Medicine, Campus Benjamin Franklin, Charité - Universitätsmedizin Berlin},
city={Berlin},
country={Germany}}

\begin{abstract}

\textbf{Background:} Precise, automated segmentation of the aorta in 4D flow MRI is essential for reproducible hemodynamic quantification. Segmentation with neural networks remains challenging due to missing dense 4D annotations and high computational effort.

\textbf{Methods:} We propose a fully automated segmentation of the ascending aorta, aortic arch and proximal descending aorta in 4D flow MRI using a four-dimensional (3D+time) U-Net. A parameter-efficient hybrid 4D kernel models temporal context, and sparse 4D labels derived from existing time-resolved 2D cross-sectional expert contours and centerlines enable training on a multicenter cohort without dense 4D annotation. Training comprised 268 scans (eight centers, two vendors). Testing used an internal set (32 scans) and an external set (30 scans, post-contrast, different site, protocol and annotator). Frame-wise 3D networks and two semi-automatic references served as comparators. Agreement of peak velocity, net flow, axial and circumferential WSS and diameters with expert contours was assessed by intraclass correlation coefficients (ICC).

\textbf{Results:} Evaluated against the time-resolved cross-sectional annotations, the 4D U-Net achieved a mean Dice similarity coefficient of 0.927 internally and 0.911 externally. The frame-wise 3D U-Net achieved 0.919 and 0.847, the static PC-MRA segmentation 0.893 internally, and registration-based propagation 0.808 externally. Differences between methods were small during systole but pronounced in low-flow diastolic phases. Agreement with the expert contours was excellent for all hemodynamic parameters (ICC $\geq$ 0.954 internally, $\geq$ 0.980 externally). Both semi-automatic references showed lower agreement.

\textbf{Conclusion:} Our method provides reproducible, time-resolved aortic segmentation, enabling automated hemodynamic analysis. It generalizes to an internal multicenter, multivendor test set and to an independent post-contrast single-center cohort. The trained model is publicly available.

\end{abstract}

\begin{graphicalabstract}
\includegraphics[width=1\linewidth]{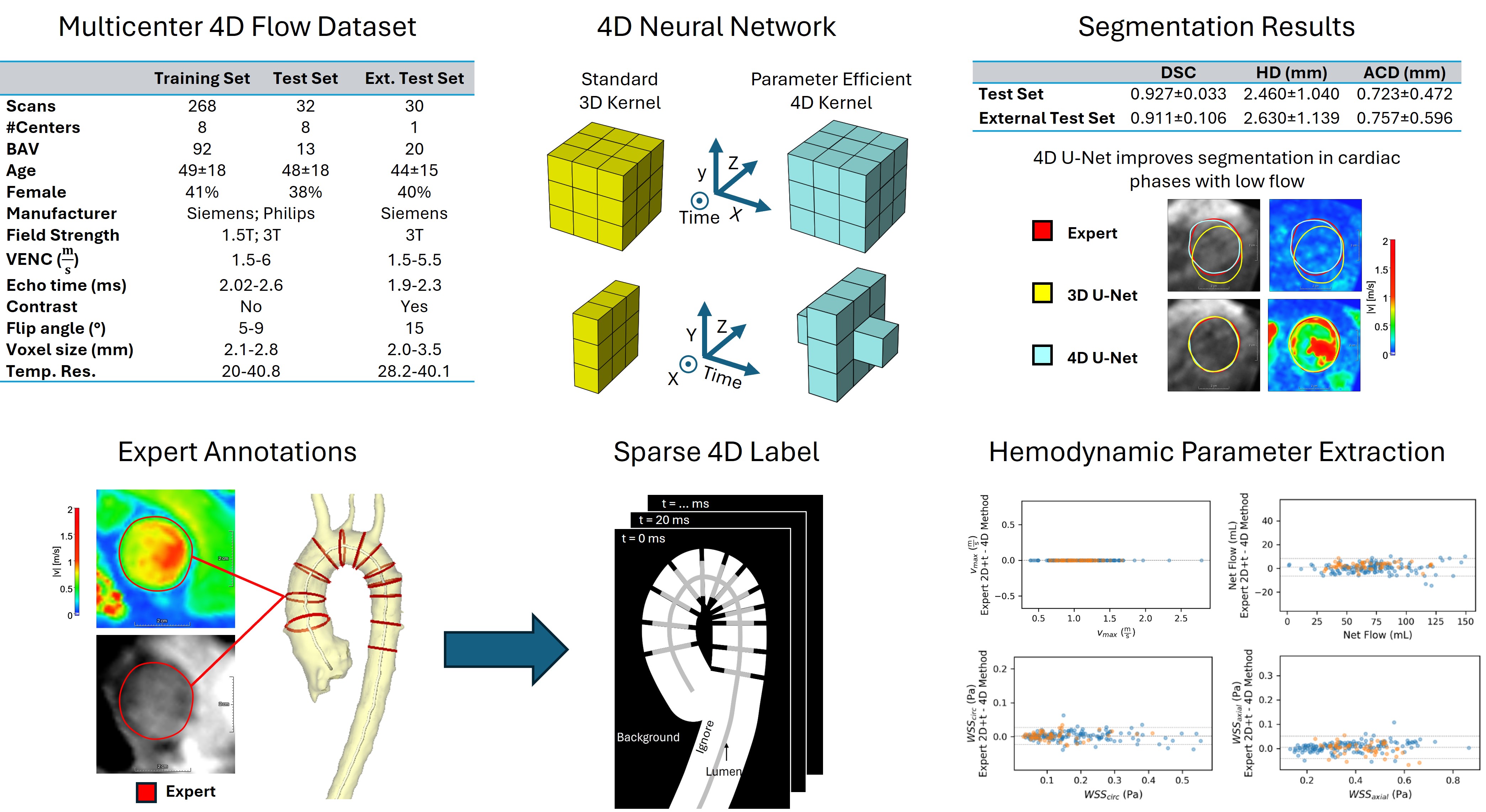}
\end{graphicalabstract}

\begin{highlights}
\item 4D convolutional neural network with hybrid kernel segments the aorta in 4D flow MRI
\item Trained on a multicenter dataset with sparse 2D+t expert annotations
\item Generalizes to an external post-contrast cohort
\item Model achieves excellent agreement on quantitative parameter extraction
\item Trained model is publicly available
\end{highlights}


\end{frontmatter}

\section{Introduction}

Four-dimensional flow magnetic resonance imaging (4D flow MRI) enables time-resolved quantification and visualization of 3D blood velocity fields using phase-contrast (PC)-MRI. Fluid-structure interactions such as wall shear stress (WSS) can be estimated using a vessel wall segmentation and assumptions about blood viscosity \cite{markl20124d}. Quantitative parameters can be used to assess pathologies in various anatomies with the main focus on valvular, aortic and congenital heart disease \cite{bissell20234d}. As an example, 4D flow MRI-derived WSS has been proposed to serve as a parameter for monitoring disease progression in patients with bicuspid aortic valve (BAV) disease \cite{barker2012bicuspid, farag2018aortic, meierhofer2013wall, lenz20204d}.

\begin{figure}
    \centering
    \includegraphics[width=0.5\linewidth]{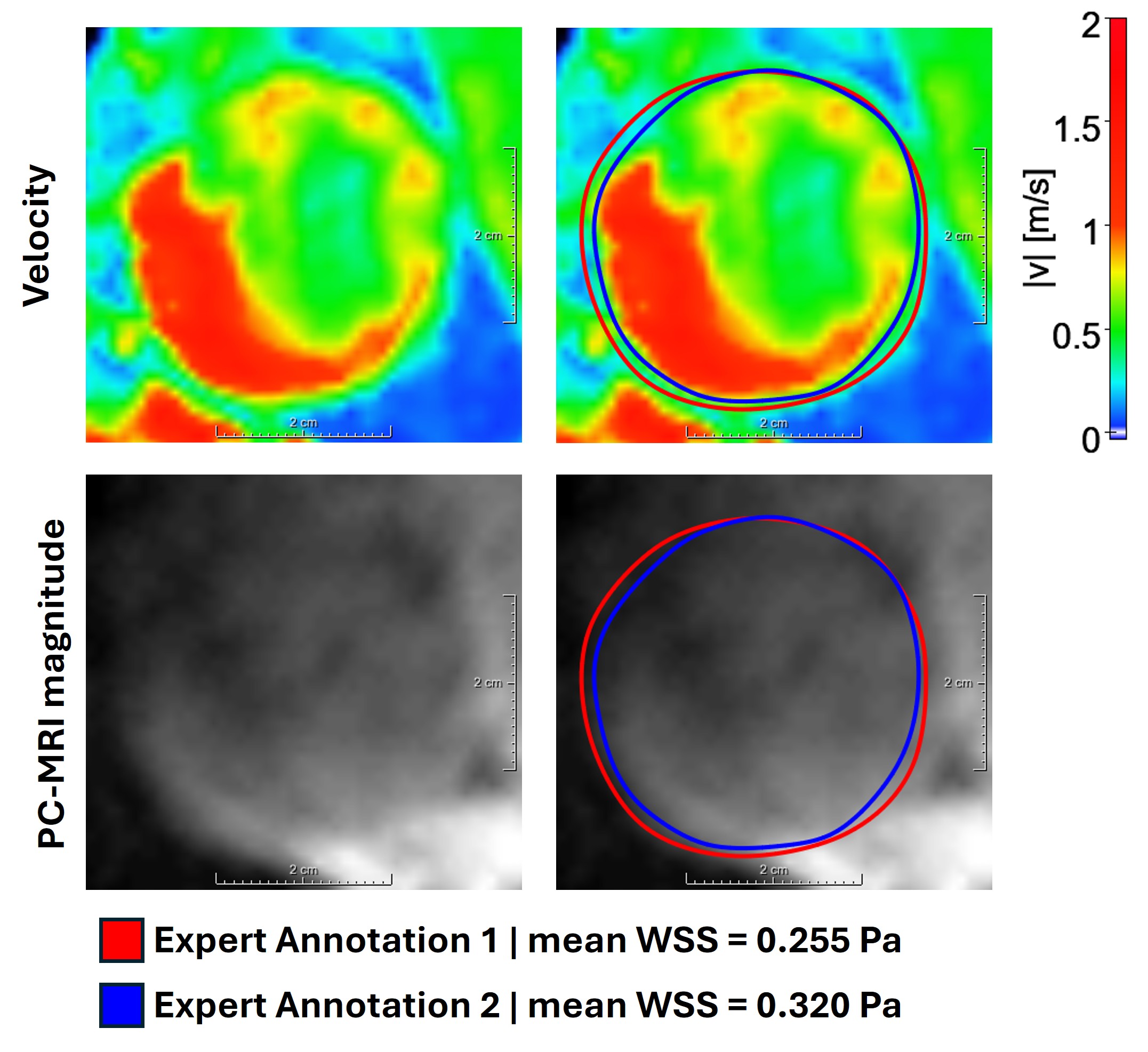}
    \caption{Illustrative example of the segmentation sensitivity of WSS. 4D flow MRI of a subject with bicuspid aortic valve with two segmentations of one cross-section in the ascending aorta. WSS was estimated assuming a constant blood viscosity of \SI{3.2}{\milli\pascal\second}. A minor difference in segmentation increases the mean WSS by 25.5\%.}
    \label{fig:introduction}
\end{figure}

These parameters are highly sensitive to the segmentation. A dynamic segmentation substantially changes WSS compared with a static one \cite{zimmermann2018wall}, and errors of less than one voxel already have a notable influence \cite{petersson2012assessment}. Figure~\ref{fig:introduction} shows an example in which a maximum contour difference of \SI{1.71}{\milli\meter} increases the mean WSS by 25.5\%. Manual and semi-automatic segmentation of dynamic 3D vascular structures is, however, labor-intensive, so clinical and research workflows typically rely on a single static 3D segmentation or on dynamic 2D cross-sections orthogonal to the vessel centerline \cite{bissell20234d}, with the associated inter- and intra-observer variability and limited scalability. An automatic segmentation is deterministic and therefore reproducible, which removes inter- and intra-observer variability.

Automated 4D aortic segmentation faces two interrelated challenges: the creation of sufficient high-quality training data, and the design of methods capable of processing 4D inputs.

\begin{figure}
    \centering
    \includegraphics[width=\linewidth]{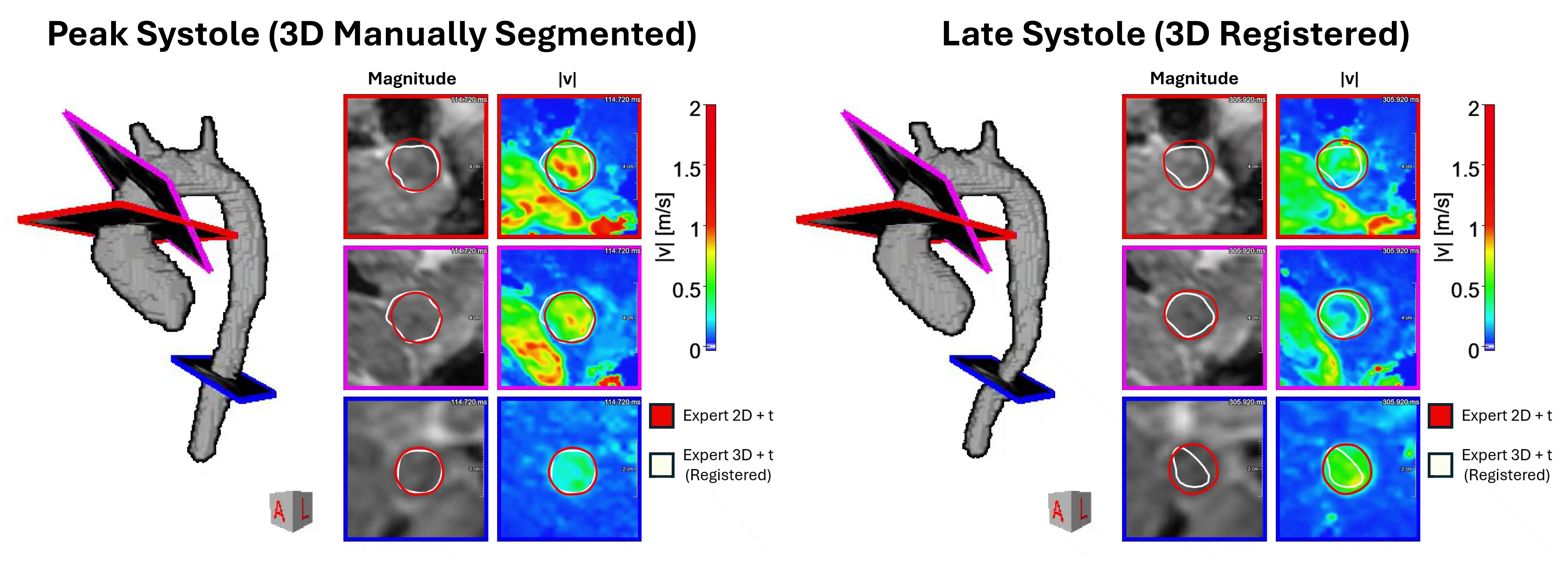}
    \caption{Qualitative example of a propagation failure. 3D aortic segmentation and three orthogonal cross-sectional views at peak systole (left) and late systole (right). At peak systole the segmentation was performed with manual oversight, at late systole it was propagated using the Morphon algorithm \cite{trenti2024oscillatory}. Manual 2D expert annotations (red) and the contour of the 3D segmentation (white) are overlaid. The propagated segmentation underestimates the lumen in the ascending aorta and loses the circular shape in the descending aorta at late systole. See supplementary video~1 for the full cardiac cycle.}
    \label{fig:propagation}
\end{figure}

\paragraph{Training data generation} Dense, time-resolved ground truth is expensive, because a full 3D segmentation must be delineated or verified in every cardiac phase (typically 20-40 frames per subject), which makes fully expert-labeled 4D datasets rare. Three existing strategies circumvent this bottleneck. Expert contours on a sparse set of time-resolved cross-sectional planes considerably reduce annotation effort and reach excellent inter-observer agreement for peak velocity \cite{manini2024impact}, but cannot be directly used to train volumetric models. Synthetic 4D geometries and flow fields \cite{garzia2023coupling, wolkerstorfer2026synthetically} provide arbitrarily large training sets, but the domain gap may limit generalization to real acquisitions with pathological anatomy and scanner-specific noise. Non-rigid registration propagates expert segmentations to the remaining cardiac phases without additional manual effort \cite{trenti2022wall, trenti2024oscillatory, knutsson2005morphons}, but relies on image intensity gradients that are weaker and less consistent in flow-encoded magnitude images than in balanced steady-state free precession (bSSFP) images. This can cause propagation failures in phases with low intravascular contrast (Figure~\ref{fig:propagation}).

\paragraph{Automatic spatio-temporal segmentation of 4D flow MRI} Early dynamic cardiac segmentation used statistical shape models \cite{perperidis2007segmentation} or an atlas-based segmentation of the systolic time point with non-rigid propagation to the remaining phases \cite{bustamante2015atlas}. Deep learning methods have been applied to the time-averaged PC-MR angiography (PC-MRA) \cite{berhane2020fully, garzia2023coupling}, which does not capture aortic motion. More recently to individual time frames, with strong results in 4D cine bSSFP \cite{merton2024assessing} and, trained on synthetic data, in 4D flow MRI \cite{wolkerstorfer2026synthetically}. Frame-independent segmentation requires sufficient contrast in every cardiac phase, which may not always be given in 4D flow MRI, where the intravascular signal in the magnitude and velocity-magnitude images is low during diastole. Temporal context from adjacent frames may compensate for this lack of evidence, which motivates a genuinely spatio-temporal model. Temporal information has been exploited for 2D+time cross-sectional segmentation \cite{manini2024impact}.

To the best of our knowledge, no prior study has used a 4D (3D+time) convolutional neural network for aortic segmentation in 4D flow MRI. The aim of this study was therefore to develop and evaluate such a network for time-resolved segmentation of the ascending aorta, aortic arch, and proximal descending aorta. We address the two challenges outlined above by implementing 4D convolutions as a sum of 3D convolutions and by deriving sparse 4D training labels from time-resolved 2D cross-sectional expert contours, which avoids the need for dense 4D annotation. We assess segmentation accuracy and the agreement of derived hemodynamic parameters against expert annotations in a multicenter cohort and in an independent external cohort acquired post-contrast, using frame-wise deep learning and semi-automatic methods as reference. The trained model and code are made publicly available.

\section{Methods}

\subsection{Data}
\label{sec:data}


\begin{table}[H]
\centering
\footnotesize
\setlength{\tabcolsep}{5.5pt}
\renewcommand{\arraystretch}{1.15}
\caption[Dataset properties]{Properties of the training, test, and external test set.
 Age and plane tilts are given as mean $\pm$ standard deviation, all other continuous
 parameters as range (min-max). The number of annotated 2D contours corresponds to the number of annotated 2D+time cross-sections multiplied by the number of reconstructed  cardiac time frames of the respective scan. BAV: bicuspid aortic valve;
 VENC: velocity encoding.}
\label{tab:dataset-properties-simple}

\begin{tabular}{lccc}
\toprule
\textbf{Characteristic}
  & \textbf{Training}
  & \textbf{Test}
  & \textbf{External test} \\
\specialrule{\lightrulewidth}{0pt}{0pt}


\rowcolor{TableSection}
\multicolumn{4}{l}{\textbf{Cohort}} \\

Scans, \(n\)
  & 268 & 32 & 30 \\

\rowcolor{TableStripe}
Age, years
  & \(49 \pm 18\)
  & \(48 \pm 18\)
  & \(44 \pm 16\) \\

Female, \%
  & 41 & 38 & 40 \\

\rowcolor{TableStripe}
BAV, \(n\)
  & 92 & 13 & 20 \\


\specialrule{\lightrulewidth}{0pt}{0pt}
\rowcolor{TableSection}
\multicolumn{4}{l}{\textbf{Image acquisition}} \\

\textbf{Manufacturer}, \(n\) 
  & & & \\

\rowcolor{TableStripe}
\quad Siemens
  & 170 
  & 22 
  & 30  \\

\quad Philips
  & 98 
  & 10 
  & 0  \\

\rowcolor{TableStripe}
\textbf{Field strength}, \(n\) 
  & & & \\

\quad \SI{3}{\tesla}
  & 180 
  & 22 
  & 30  \\

\rowcolor{TableStripe}
\quad \SI{1.5}{\tesla}
  & 88 
  & 10 
  & 0  \\

\textbf{Acquisition plane}, \(n\) 
  & & & \\

\rowcolor{TableStripe}
\quad Sagittal
  & 69 
  & 6 
  & 10  \\

\quad Oblique sagittal
  & 199 
  & 26 
  & 20  \\

\rowcolor{TableStripe}
\quad Coronal tilt, \si{\degree}
  & \(21.8 \pm 8.0\)
  & \(21.4 \pm 8.2\)
  & \(22.7 \pm 10.1\) \\

\quad Axial tilt, \si{\degree}
  & \(0.4 \pm 6.2\)
  & \(0.0 \pm 2.5\)
  & \(0.8 \pm 3.4\) \\

\rowcolor{TableStripe}
Contrast agent
  & No & No & Yes \\

VENC, \si{\meter\per\second}
  & \(1.5\text{-}6.0\)
  & \(1.5\text{-}6.0\)
  & \(1.5\text{-}5.5\) \\

\rowcolor{TableStripe}
Echo time, \si{\milli\second}
  & \(2.02\text{-}2.60\)
  & \(2.02\text{-}2.60\)
  & \(1.90\text{-}2.30\) \\

Flip angle, \si{\degree}
  & \(5\text{-}9\)
  & \(5\text{-}9\)
  & 15 \\

\rowcolor{TableStripe}
Temporal resolution, \si{\milli\second}
  & \(20.0\text{-}40.8\)
  & \(20.0\text{-}40.8\)
  & \(38.2\text{-}40.1\) \\

Acquired voxel size, \si{\milli\meter}
  & \(2.1\text{-}2.8\)
  & \(2.1\text{-}2.8\)
  & \(2.0\text{-}3.5\) \\

\rowcolor{TableStripe}
Reconstructed voxel size, \si{\milli\meter}
  & \(1.0\text{-}3.0\)
  & \(1.21\text{-}2.80\)
  & \(2.0\text{-}3.5\) \\


\specialrule{\lightrulewidth}{0pt}{0pt}
\rowcolor{TableSection}
\multicolumn{4}{l}{\textbf{Annotations}} \\

2D+time cross-sections, \(n\)
  & \num{2954}
  & \num{366}
  & \num{222} \\

\rowcolor{TableStripe}
Resulting 2D contours, \(n\)
  & \num{90354}
  & \num{10825}
  & \num{6170} \\

\bottomrule
\end{tabular}
\end{table}

\subsubsection{Study Population and Acquisitions}
As main dataset we used the 4D flow MRI data and expert annotations also used by Manini et al. \cite{manini2024impact}, excluding one subject whose consent for further analysis was revoked. In contrast to Manini et al., who separated 41 datasets with additional valve configurations and post-surgical scans, we combined all available data to obtain the largest and most heterogeneous dataset possible. The dataset covers eight imaging centers, two vendors, seven scanner models, and multiple PC-MRI acquisition protocols. It comprises 23 healthy subjects \cite{demir2022traveling, wiesemann2021impact}, 105 subjects with a BAV \cite{nordmeyer2021circulatory, hanigk2023aortic, wiesemann2023changes, lenz20204d}, 16 subjects with a stenotic tricuspid valve \cite{nordmeyer2021circulatory}, 5 subjects with a unicuspid aortic valve \cite{lenz20204d}, and 131 subjects from the general population without known aortic pathology (who may present with other cardiovascular conditions). Post-surgical scans were obtained for 20 subjects \cite{lenz20204d}, resulting in 300 MRI scans from 280 subjects.

The volumes were acquired sagittal or oblique sagittal to efficiently cover the ascending aorta, the aortic arch, and the thoracic descending aorta. For the oblique acquisitions, the stated anterior-posterior (AP) and left-right (LR) axis values refer to the oblique axes. The acquired field of view (FoV) and the reconstructed voxel sizes varied considerably across the multicenter cohort (Figure~\ref{fig:scan_parameters}). All acquisitions covered the aortic root and left ventricular outflow tract, the ascending aorta, and the aortic arch, whereas coverage of the descending aorta varied between sites, ranging from the level of the basal cardiac chambers to the aortic bifurcation. The annotated target region (ascending aorta, aortic arch, and proximal descending aorta; Figure~\ref{fig:annotations}) was located entirely within the field of view in all datasets. Image preprocessing included background phase offset correction and phase unwrapping.

\subsubsection{Data Split}
\begin{figure}
    \centering
    \includegraphics[width=1\linewidth]{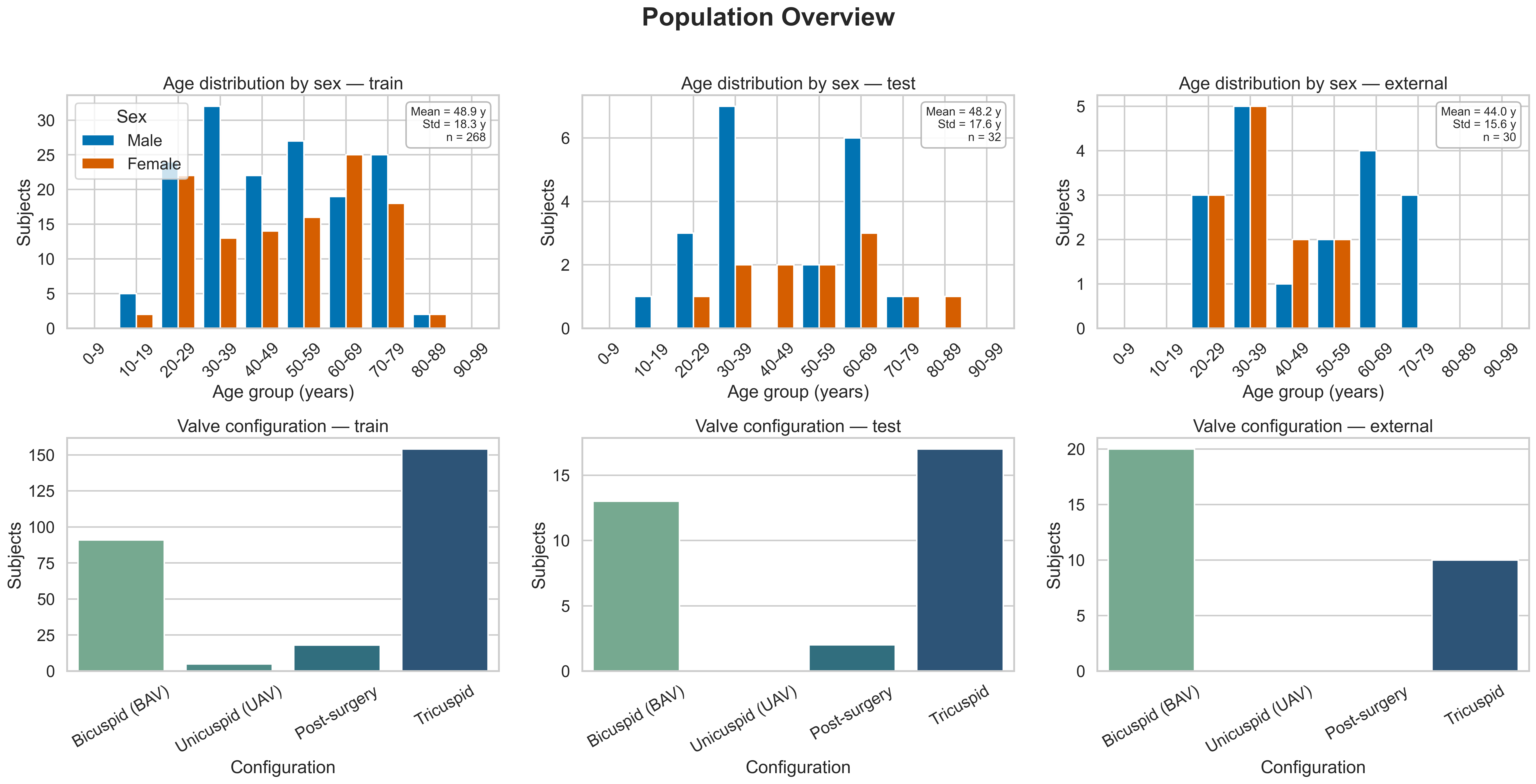}
    \caption{Study population in the training set, test set, and external test set.}
    \label{fig:population}
\end{figure}

\begin{figure}
    \centering
    \includegraphics[width=0.7\linewidth]{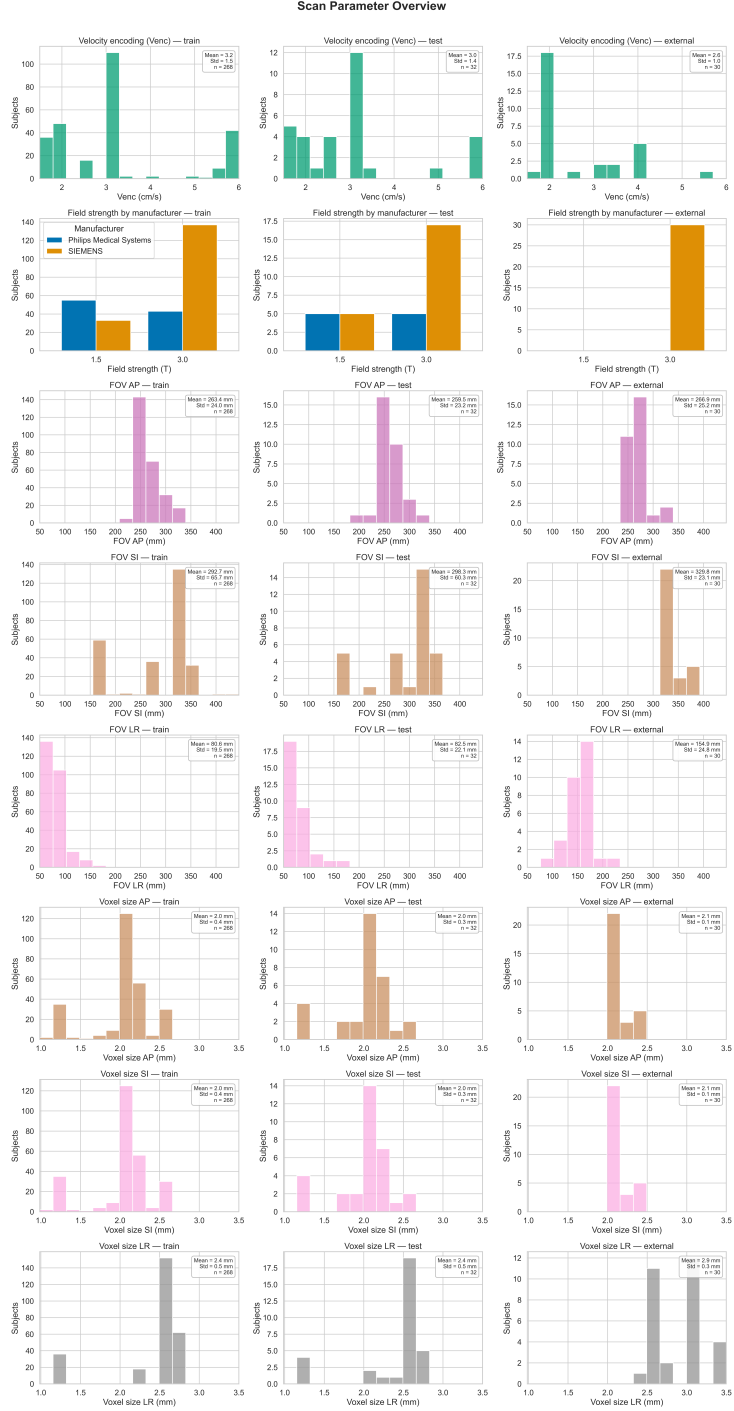}
    \caption{Scan parameters of the training set, test set, and external test set. The FoV and reconstructed voxel size are stated in anterior-posterior (AP), left-right (LR), and superior-inferior (SI) direction.}
    \label{fig:scan_parameters}
\end{figure}

We split the data at the subject level. The test set contains 30 subjects, two of whom also had a post-surgical examination, resulting in 32 MRI datasets. The training set contains the remaining 250 subjects (268 MRI scans), on which we performed subject-wise 5-fold cross-validation for all method development, i.e.\ for all configuration decisions described in Section~\ref{sec:selection}. Table~\ref{tab:dataset-properties-simple} and Figures~\ref{fig:population} and~\ref{fig:scan_parameters} describe the dataset characteristics, with statistics reported per scan rather than per subject, so that subjects with a pre- and post-surgical scan contribute their characteristics twice.

\subsubsection{External Test Set}
\label{sec:ets}
We evaluated model generalization on an external test set of 20 patients with BAV and 10 healthy subjects. Unlike the training data, scans were acquired 5-10 minutes post-contrast and with a higher flip angle of \SI{15}{\degree} \cite{trenti2024oscillatory}. Demographics and scan parameters are reported in Table~\ref{tab:dataset-properties-simple}, Figure~\ref{fig:population}, and Figure~\ref{fig:scan_parameters}. To assess robustness to variations in cross-sectional position, we sampled cross-sections every \SI{20}{\milli\meter} from the aortic annulus, beyond the left subclavian artery, with additional slices in the proximal and distal descending aorta.

\subsection{Reference Annotations}
\label{sec:annotation}

All annotations used for network training existed prior to this work and were already used for a 2D+time segmentation \cite{manini2024impact}.

\subsubsection{3D PC-MRA Segmentation and Centerline Definition}
\label{sec:annotation3d}
A 3D aortic segmentation was created on the time-averaged PC-MRA using the watershed algorithm with manual corrections. The static aortic centerline was derived from it using distance-transform-based skeletonization \cite{selle2002analysis} with expert oversight and manual correction where necessary.

The PC-MRA was computed from the magnitude images $M(t)$ and the velocity magnitude $|\mathbf{v}(t)|=\sqrt{v_x^2+v_y^2+v_z^2}$ as $\mathrm{PCMRA}=\frac{1}{T}\sum_{t=1}^{T}{M}(t)^{2}\,|\mathbf{v}(t)|^{2}$, averaged over all $T$ reconstructed cardiac phases.

\subsubsection{2D+time Cross-Sectional Annotations}
\label{sec:annotation2dt}
\begin{figure}
    \centering
    \includegraphics[width=1\linewidth]{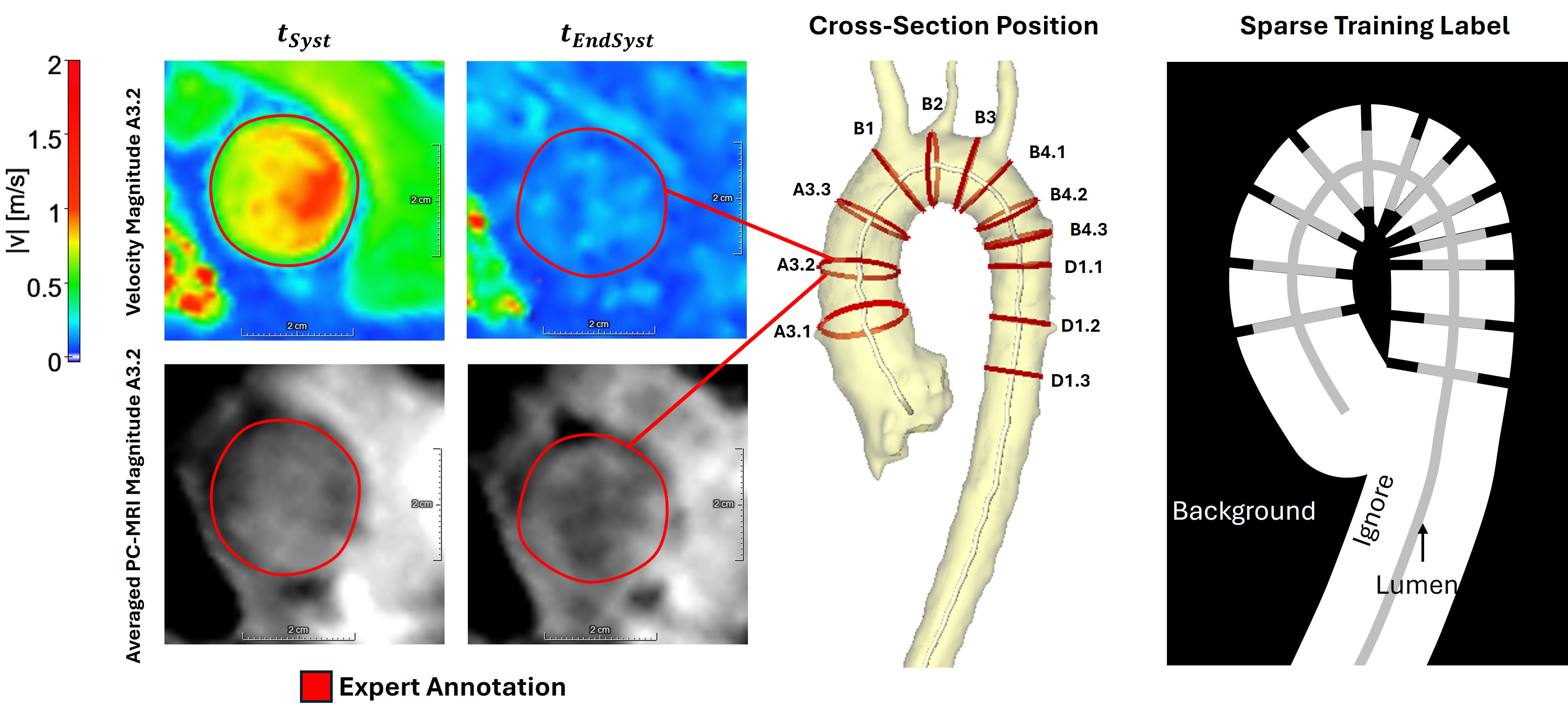}
    \caption{Sparse cross-section annotations and label for vessel segmentation in 4D flow MRI. On the left, the velocity magnitude and the averaged PC-MRI magnitude are shown for the systolic and the end-systolic time points. The expert annotation was created for all time points and is shown in red. In the middle, the twelve cross-section positions are shown, spanning the ascending aorta, aortic arch, and proximal descending aorta. The right part illustrates a schematic of the corresponding sparse label: background (black) and lumen (gray) voxels are used to compute the loss during network training, while ignore voxels (shown in white) are excluded from the loss calculation.}
    \label{fig:annotations}
\end{figure}

Twelve cross-sectional planes were placed perpendicular to the centerline at the predefined anatomical locations introduced by Schafstedde et al.\ \cite{schafstedde2023population} (Figure~\ref{fig:annotations}, middle). Six of them were placed manually: A3.1 at the sinotubular junction; B1 and B2 before and after the brachiocephalic trunk; B3 and B4.1 before and after the left subclavian artery; and D1.1 in the descending aorta at the level of the pulmonary artery. The remaining planes were placed automatically: A3.2 and A3.3 equidistant between A3.1 and B1; B4.2 and B4.3 equidistant between B4.1 and D1.1; and D1.2 and D1.3 at the same distance to D1.1 as used for A3.2 and A3.3 relative to A3.1 \cite{schafstedde2023population}.

Along these planes, a time-resolved multiplanar reconstruction (MPR) of the magnitude image (averaged over the four flow encodings) and of the velocity magnitude image was created using trilinear interpolation (Figure~\ref{fig:annotations}, left). Vessel lumen contours were manually created in these MPRs across all cardiac time frames by trained experts. Complete annotation details are provided in the original publication \cite{manini2024impact}.

\subsubsection{4D Volumetric Expert Annotation}
\label{sec:annotation4d}
To relate the cross-sectional evaluation to a volumetric one (Section~\ref{sec:validation2dt}), a different expert created 4D annotations for six cases of the internal test set: the cases with the minimal, median, and maximal DSC of the proposed method and three randomly selected cases. This selection deliberately spans the full range of segmentation quality and is not representative, so the absolute metric values on this subset must not be interpreted as performance estimates. The expert drew 2D+time contours at \SI{10}{\milli\meter} intervals along the centerline, from the sinotubular junction to the distal end of the manual PC-MRA segmentation, and the per-frame contours were converted to masks by Poisson surface reconstruction \cite{kazhdan2006poisson}, which reaches substantially higher inter-observer agreement than volumetric segmentation with a 3D brush tool \cite{brosig2024learning}.

\subsection{Automatic 4D Aortic Segmentation}
\label{sec:pipeline}

The proposed pipeline consists of two steps. First, the aorta is localized by a 3D segmentation of the time-averaged PC-MRA, which is used solely to define the region of interest (ROI). Second, a 4D CNN segments the aorta over the complete cardiac cycle within this ROI. The resulting 4D segmentation is transformed back into the original image space.

\subsubsection{Region of Interest Localization and Cropping}
\label{sec:cropping}
To reduce training and inference runtime, the 4D volume was cropped to the bounding box of a 3D aortic mask of the time-averaged PC-MRA, dilated by a fixed 5-voxel margin so that the lumen remains contained despite minor localization errors. During training, the manual 3D PC-MRA segmentation (Section~\ref{sec:annotation3d}) was used. At test time, the aorta was localized automatically by a localizer 3D nnU-Net (default \texttt{3d\_fullres}, 5-fold ensemble) trained on the internal training set with the manual PC-MRA segmentations as labels. Details of the localizer, of the PC-MRA computation, its cross-validation performance, and its performance on the test sets are given in~\ref{app:localizer}.

\subsubsection{Network Input, Preprocessing, Temporal Resampling, and Patching}
\label{subsec:preprocessing}
The network receives two input channels: the PC-MRI magnitude image, averaged over the four flow encodings, and the velocity magnitude image $|\mathbf{v}(t)| = \sqrt{v_x(t)^2 + v_y(t)^2 + v_z(t)^2}$. Spatially, we applied the standard nnU-Net resampling strategy to the median in-plane resolution of \SI{2.125}{\milli\meter} and the median slice spacing of \SI{2.5}{\milli\meter}. Temporally, each dataset was resampled to 32 time frames to account for the differing temporal sampling rates across sites. Following nnU-Net's approach, images were resampled with cubic spline interpolation and labels as one-hot encoded class maps, preserving the discrete classes including the ignore label \cite{isensee2021nnu}, and the resampling was inverted after inference. For training and inference we used spatial patches of $32 \times 64 \times 64$ voxels covering all 32 time frames.

\subsubsection{4D Convolution and Convolution Kernels}
\label{sec:conv4d}

The inputs are 4D volumes of size $T \times Z \times Y \times X$. We therefore consider 4D convolutions of the general form

\begin{equation}
y[c_{\text{out}}, p] = b[c_{\text{out}}] +
\sum_{c_{\text{in}}=0}^{C_{\text{in}}-1} \; \sum_{\Delta \in \mathcal{S}}
W\bigl[c_{\text{out}}, c_{\text{in}}, \Delta\bigr] \cdot
\mathbf{x}\bigl[c_{\text{in}}, \; s \odot p + \Delta\bigr],
\label{eq:conv4d}
\end{equation}

where $p = (t, z, y, x)$ is the output position, $s$ the stride, $\odot$ the element-wise product, and $\mathcal{S} \subseteq \{-1,0,1\}^4$ the kernel support, i.e.\ the set of temporal and spatial offsets $\Delta = (\Delta t, \Delta z, \Delta y, \Delta x)$ relative to the kernel center $\Delta = (0,0,0,0)$. The hypercube kernel uses the full support $\mathcal{S}_{\text{hyper}} = \{-1,0,1\}^4$, i.e.\ $81$ weights per input-output channel pair, and is realized as a sum of three 3D convolutions following Myronenko et al.\ \cite{myronenko20194d} (\ref{app:detailed_4d_conv}).

Following an approach that has proven successful for sparse convolutions \cite{choy20194d}, but that to our knowledge has not yet been used in dense 4D CNNs, we define the hybrid kernel support

\begin{equation}
\mathcal{S}_{\text{hybrid}} = \bigl\{\, \Delta \in \mathcal{S}_{\text{hyper}}
\;:\; \Delta t = 0 \;\lor\; (\Delta z, \Delta y, \Delta x) = (0,0,0) \,\bigr\},
\label{eq:hybrid}
\end{equation}

i.e.\ a full $3 \times 3 \times 3$ spatial kernel on the center frame and a single $1 \times 1 \times 1$ weight on each of the two adjacent frames, requiring only $29 = 27 + 2$ weights per channel pair. Unless stated otherwise, all reported results use the hybrid kernel.

\paragraph{Padding}
All 4D convolutional layers use zero padding in the spatial dimensions and cyclic padding in the temporal dimension, i.e.\ $\mathbf{x}[t=-1] = \mathbf{x}[t=T-1]$ and $\mathbf{x}[t=T] = \mathbf{x}[t=0]$. This reflects the underlying physiology, where the last reconstructed frame is temporally adjacent to the first and aortic geometry and flow are approximately periodic.

\subsubsection{Network Architecture}
\begin{figure}
    \centering
    \includegraphics[width=1\linewidth]{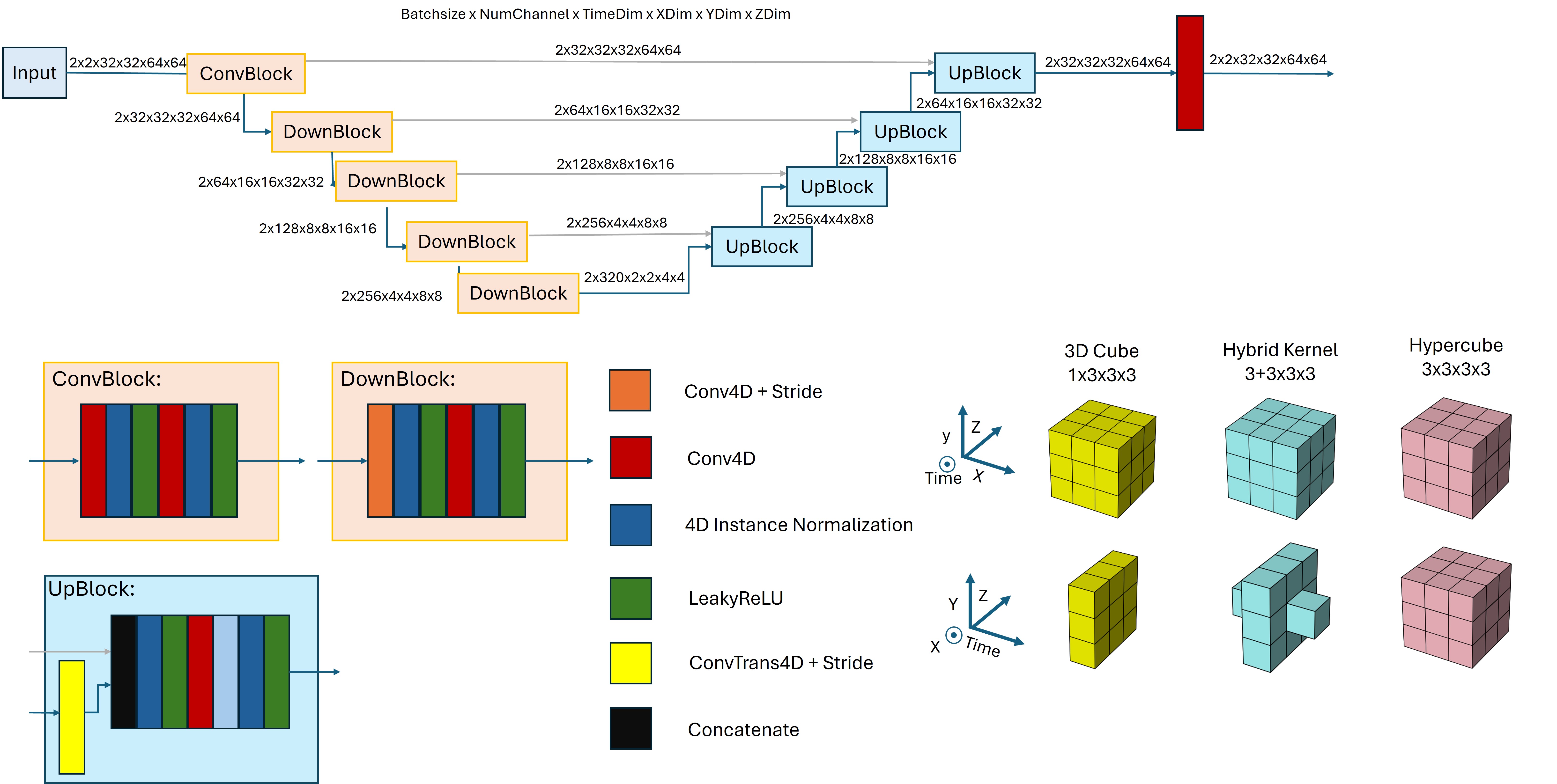}
    \caption{Network architecture of the proposed 4D U-Net. The Conv4D is performed either using the hybrid kernel or the hypercube kernel. The 3D cube, which is common for 3D CNNs, is shown for reference.}
    \label{fig:network_architecture}
\end{figure}

The network is based on the U-Net architecture. It receives the two input channels described in Section~\ref{subsec:preprocessing} and outputs a binary mask, with zero representing background and one representing the aortic lumen. The implementation follows the architecture used in the nnU-Net framework \cite{isensee2021nnu}, extended to a fourth dimension by replacing all 3D convolutions with the 4D convolutions of Section~\ref{sec:conv4d}. The architecture and the kernel shapes are shown in Figure~\ref{fig:network_architecture}.

\subsubsection{Sparse Training Labels}
\label{sec:sparse_labels}
The 2D+time cross-sectional annotations were transformed into sparse 4D labels (Figure~\ref{fig:annotations}, right) by applying the method of Rahlfs et al.\ \cite{rahlfs5564923learning} per time frame. Voxels closer to the centerline than the centerline radius (cr) are labeled lumen, voxels farther away than the background radius are labeled background, and voxels within half the extrusion thickness (et) of a cross-sectional annotation plane are projected onto that plane and labeled lumen if the projection falls inside the annotated contour and background otherwise. All remaining voxels, and all voxels for which two rules disagree, are assigned the ignore label and do not contribute to the loss. The per-frame labels are concatenated along the temporal axis. The centerline-derived labels are static, whereas the cross-sectional labels vary over time. A formal definition is given in \ref{app:sparse_labels}.

The background radius was fixed to \SI{41.9}{\milli\meter}. The centerline radius and the extrusion thickness were treated as free parameters and selected on the training set (Section~\ref{sec:selection}).

\subsubsection{Data Augmentation and Training}
\label{sec:training}
We employed all data augmentations used by nnU-Net and extended them to 4D, i.e.\ a transform with the same parameters was applied to all time frames of a training sample. For training with sparse annotations we used the masked Dice and masked cross-entropy loss proposed by Gotkowski et al.\ \cite{gotkowski2025revisiting}, with both terms weighted equally as in the nnU-Net default. Voxels carrying the ignore label were excluded from both. Optimization used stochastic gradient descent, Nesterov momentum 0.99, an initial learning rate of 0.001, nnU-Net learning-rate decay with a batch size of 2 and 250 mini-batches per epoch. In deviation from the nnU-Net default of 1000 epochs, all models were trained for 500 epochs to keep the total training time feasible, using the same budget for all compared configurations, in a subject-wise 5-fold cross-validation on the internal training set.

\subsubsection{Inference}
Inference followed the standard nnU-Net procedure. The five cross-validation models were applied as an ensemble by averaging their softmax outputs, using sliding-window inference with a tile step size of 0.5, Gaussian tile weighting, and mirroring-based test-time augmentation on the spatial axes \cite{isensee2021nnu}. The final label map was obtained by taking the argmax over the class probabilities. No postprocessing such as connected-component filtering was applied. The predicted 4D segmentation was resampled back to the original temporal resolution and written into the original image space, so that all reported metrics are computed in the original image geometry.

\subsubsection{Training and Inference Time Measurement}
We trained and evaluated all models on NVIDIA H200 80\,GB GPUs. Training time is reported for fold~0 of each configuration. Inference time was assessed on the 32 internal test set cases using the five-model ensemble, includes data and model loading, and is reported as the average per case. The processing time for PC-MRA computation, ROI localization, cropping, and back-transformation does not depend on the segmentation model and is reported separately.

\subsection{Model and Label Parameter Selection}
\label{sec:selection}

Model configuration and sparse-label parameters were selected on the 5-fold cross-validation of the internal training set, using the mean DSC over all annotated 2D cross-sections and time frames as the only selection criterion. Because both interact, they were optimized with a coordinate-descent strategy over two axes: the model configuration (the original 3D nnU-Net \cite{isensee2021nnu} applied independently to each time frame, the 4D hypercube kernel, the 4D hypercube kernel with one third of the feature maps per layer to approximately match the parameter count of the hybrid kernel, and the 4D hybrid kernel with and without temporal resampling to 32 frames of Section~\ref{subsec:preprocessing}) and the sparse-label creation with $\text{cr}\in\{2,4\}~\si{\milli\meter}$ and $\text{et}\in\{2,4,6\}~\si{\milli\meter}$. The procedure was iterated until neither axis improved the mean DSC; the resulting configuration 4D hybrid kernel with temporal resampling and cr = 4 mm, et = 6 mm was used for all evaluations on the test set and the external test set. All steps and all evaluated configurations are reported in \ref{app:coord}.

\subsection{Comparison Methods}
\label{sec:comparison_methods}

We compared the proposed method (nnU4D-sp) against six baseline segmentations. All learning-based baselines with the ending -sp used the same training set, sparse labels, subject-wise 5-fold cross-validation, augmentation, loss, optimizer, batch sizes ($\text{batch sizes}\times \text{timeframes}$ for 3D methods) as nnU4D-sp.

\paragraph{nnU3D-sp} A 3D nnU-Net applied independently to each time frame. This corresponds to the $3\times3\times3$ kernel configuration of the ablation study. nnU-Net serves as a strong out-of-the-box baseline for 3D segmentations \cite{isensee2021nnu}.

\paragraph{UTR3D-sp} The hybrid convolution/transformer network UNETR \cite{hatamizadeh2022unetr}, applied independently to each time frame. UNETR did not undergo any architecture- or task-specific hyperparameter tuning.

\paragraph{nnU3D-syn} A 3D nnU-Net that uses only the magnitude data as input and was trained on synthetic data of 50 synthetic subjects generated from 28 subjects with a mean age of 75 years \cite{wolkerstorfer2026synthetically}. The trained model was taken unchanged from the repository of Wolkerstorfer et al.\ \cite{wolkerstorfer2026synthetically}, performing the required resampling and transformations as described there, and applied independently to each time frame.

\paragraph{nnU3D-syn+} An extended version of nnU3D-syn, trained with additional datasets and a label definition that includes the aortic root. The trained model was taken unchanged from the repository of Wolkerstorfer et al.\ \cite{wolkerstorfer2026synthetically} and applied independently to each time frame.

\paragraph{PCMRA3D} The semi-automatic 3D PC-MRA segmentation described in Section~\ref{sec:annotation3d}, applied statically to all time points. It was evaluated only on the internal test set, as no PC-MRA segmentation created with the same workflow was available for the external test set.

\paragraph{ManReg4D} The semi-automatic approach proposed by Trenti et al.\ \cite{trenti2022wall}, which generates a 3D segmentation of the aorta at the systolic time point in a semi-automated workflow and propagates it to the remaining time frames using the Morphon algorithm. We used the annotations created for the prior publication \cite{trenti2024oscillatory}. Therefore, it was evaluated only on the external test set.

\subsection{Evaluation}
\label{sec:evaluation}

\subsubsection{Evaluation on 2D+time Cross-Sectional Contours}
\label{sec:eval2dt}
\begin{figure}
    \centering
    \includegraphics[width=0.7\linewidth]{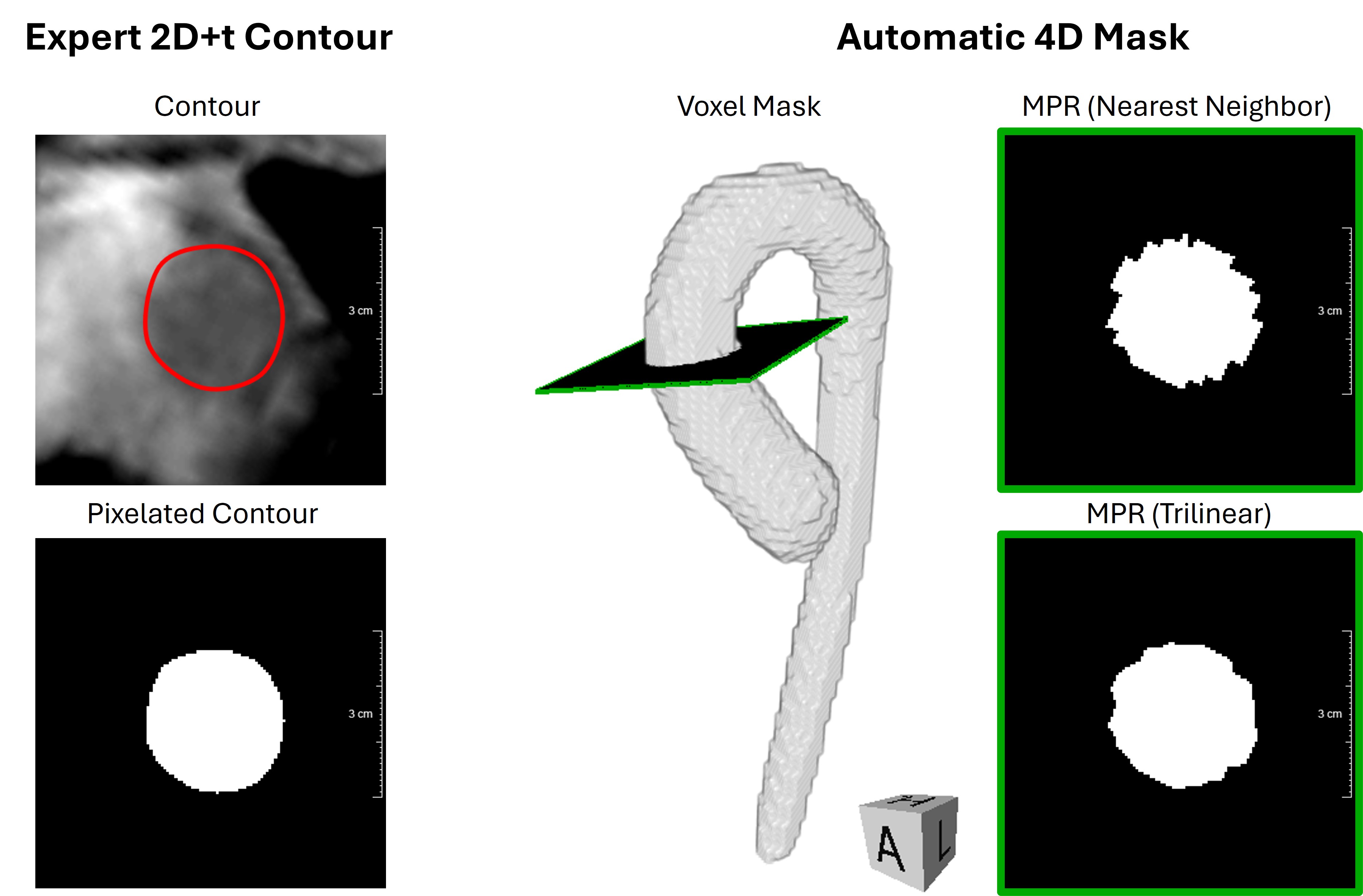}
    \caption{Transformations used to evaluate automatic 4D segmentation against time-resolved 2D contours. Left: pixelated expert annotation shown within the annotation plane. Right: a single time point of the automatic 4D label with the corresponding annotation plane overlaid. A multiplanar reconstruction (MPR) is extracted along this plane for model evaluation. Trilinear interpolation with thresholding at 0.5 is applied to produce smoother label boundaries compared to nearest-neighbor interpolation.}
    \label{fig:Evalution_Cross_Section}
\end{figure}

All methods were evaluated against the time-resolved cross-sectional expert annotations of the internal and the external test set. Figure~\ref{fig:Evalution_Cross_Section} shows the pixelated expert contour and the MPR along the annotation plane extracted from the automatic 4D segmentation mask. It was computed at an isotropic resolution of \SI{0.5}{\milli\meter} and a field of view of \SI{64}{\milli\meter}. We used trilinear interpolation and a threshold of 0.5, which yields a smoother boundary than nearest-neighbor interpolation.

For each annotated cross-section and time point we computed the Dice similarity coefficient (DSC), the Hausdorff distance (HD), and the average contour distance (ACD). Cross-sections and time points for which a method predicted no lumen are reported as failure rate. Failures entered the DSC with DSC~=~0 and were excluded from HD and ACD, because both distances are undefined in the absence of a predicted contour. Distance metrics must therefore be interpreted together with the failure rate. We additionally report all metrics per cross-sectional position.

\subsubsection{Segmentation Performance Across Cardiac Phases}
To assess the influence of the cardiac phase, we plotted the mean DSC and the area between the 5th and 95th percentiles over time. To account for the variable temporal sampling rates and cycle durations, all curves were interpolated onto a uniform grid with \SI{8}{\milli\second} sampling over an \SI{800}{\milli\second} cardiac cycle.

\subsubsection{Quantitative Parameter Extraction}
\label{sec:quantparam}
Quantitative parameters were computed for each annotated 2D+time cross-section, using the plane normal $\mathbf{n}$, the cross-sectional area segmented as lumen $A$, its boundary contour $C$, and the contour tangent $\boldsymbol{\tau}$. Wall shear stress (WSS) was estimated at the segmented lumen boundary from the through-plane and in-plane velocity gradients, assuming Newtonian behavior with a dynamic blood viscosity of \SI{0.0032}{\pascal\second}. The velocity gradient was evaluated directly at the segmented contour, i.e.\ without an inward offset along the contour normal, and no spatial or temporal smoothing was applied.

The maximum velocity over the cardiac cycle was computed as the maximum through-plane velocity over all lumen voxels and all time frames:
\begin{equation}
v_{\max} = \max_{t \in T} \max_{(x,y,z)\in A(t)} \mathbf{v} (x,y,z,t)\cdot \mathbf{n}
\end{equation}

The net flow over the cardiac cycle:
\begin{equation}
\text{Net Flow} = \int_{t\in T} \int_{(x,y,z)\in A(t)} \mathbf{v}(x,y,z,t)\cdot \mathbf{n} \; dA \, dt
\end{equation}

The average axial WSS at the time frame with maximal average axial WSS:
\begin{equation}
WSS_{\text{axial}} = \max_{t\in T} \; \frac{1}{\lvert C(t)\rvert} \int_{(x,y,z)\in C(t)} \mathbf{WSS} (x,y,z,t)\cdot \mathbf{n} \; dc
\end{equation}

The average circumferential WSS at the time frame with maximal average circumferential WSS:
\begin{equation}
WSS_{\text{circ}} = \max_{t\in T} \; \frac{1}{\lvert C(t)\rvert} \int_{(x,y,z)\in C(t)} \mathbf{WSS} (x,y,z,t)\cdot \boldsymbol{\tau}(x,y,z,t) \; dc
\end{equation}

The systolic time point $t_{\text{Syst}}$ was defined as the time frame with maximum net flow through the expert contour, and the end-systolic time point $t_{\text{EndSyst}}$ as the first time frame after $t_{\text{Syst}}$ at which the net flow no longer decreases. Both were determined once from the expert annotation and used identically for all methods. For these two time points we computed the area-based diameter
\begin{equation}
D(t) = 2 \sqrt{ \frac{ \int_{(x,y,z)\in A(t)} dA }{\pi} }
\end{equation}

Quantitative parameters are reported for the three methods with the lowest failure rate of the respective test set. For comparability, they were computed only on the cross-sections successfully segmented by all of these methods, which may introduce a positive bias.

\subsubsection{Validation of the 2D+time Evaluation Against 4D Volumetric Evaluation}
\label{sec:validation2dt}
To assess whether the cross-sectional metrics support conclusions comparable to a fully volumetric evaluation, we used the six internal test cases with both 2D+time and 4D expert annotations (Section~\ref{sec:annotation4d}) and all methods of Section~\ref{sec:comparison_methods}, i.e.\ 36 automatic segmentations spanning a wide range of segmentation quality. Per case and method we compared $\text{mean}(\text{DSC}_{\text{2Dt}})$ with $\text{DSC}_{\text{4D}}$, $\text{mean}(\text{ACD}_{\text{2Dt}})$ with the average surface distance $\text{ASD}_{\text{4D}}$, and $\max(\text{HD}_{\text{2Dt}})$ with the 95th-percentile Hausdorff distance $\text{HD95}_{\text{4D}}$, where the volumetric metrics were computed per time frame against the 4D expert annotation and averaged over all time frames.

\subsubsection{Statistical Analysis}
\label{sec:statistics}
Segmentation metrics were computed per annotated 2D cross-section and time frame and are reported as mean $\pm$ standard deviation over all cross-sections and time frames. Quantitative parameters were computed per annotated 2D+time cross-section and the validation of the 2D+time against 4D metrics on case-wise values ($n = 36$ case-method combinations).

Agreement was assessed with Bland-Altman analysis, reporting bias and 1.96 standard deviations of the differences, and with the intraclass correlation coefficient ICC(2,1) (two-way random effects, single measurement, absolute agreement) with its 95\% confidence interval. Following Koo and Li \cite{koo2016guideline}, ICC values $\geq$~0.90 were considered excellent, 0.75-0.90 good, 0.50-0.75 moderate, and $<$~0.50 poor. Linear association was additionally quantified with Pearson's correlation coefficient ($r$) and agreement of the resulting method ranking with Spearman's rank correlation coefficient ($\rho$).

\section{Results}

\subsection{Agreement Between 2D+time Cross-Sectional and 4D Evaluation Metrics}

Case-wise 2D+time and 4D volumetric metrics agreed excellently for DSC (ICC 0.985, bias $+0.012$) and moderate for $\text{mean}(\text{ACD}_{\text{2Dt}})$ versus $\text{ASD}_{\text{4D}}$ (ICC 0.741, bias $-0.15$~\si{\milli\meter}), whereas $\max(\text{HD}_{\text{2Dt}})$ was not numerically interchangeable with $\text{HD95}_{\text{4D}}$ (ICC 0.156, bias $-7.06$~\si{\milli\meter}) despite a monotonic association ($\rho = 0.780$). All subsequent results are therefore reported with 2D+time metrics, using DSC and the average-distance measure as approximate proxies for their volumetric counterparts and the 2D+time HD as a secondary relative measure only. A detailed analysis of the differences between 2D+time and 4D metrics can be found in \ref{app:2dt_vs_4d}.

\subsection{Ablation Studies}

All 4D kernel shapes reached a higher DSC and a lower HD and ACD than the frame-wise 3D kernel (DSC 0.9236-0.9251 vs.\ 0.9157), while differences between the 4D variants were within 0.0015 DSC. The hybrid kernel with temporal resampling and the hypercube kernel reached near identical highest mean DSC (hybrid: 0.9251, hypercube: 0.9250). With 29 instead of 81 weights per channel combination, the hybrid kernel had a 27\% shorter training time and a 30\% shorter inference time than the hypercube kernel with the full channel count (Table~\ref{tab:ablation}). For the sparse labels, et~=~\SI{6}{\milli\meter} with cr~=~\SI{4}{\milli\meter} achieved highest DSC and clearly lowest failure rate (Table~\ref{tab:ablation_sampling}). All configurations of the four coordinate-descent steps are reported in \ref{app:coord}.

\begin{table}[t]
\centering
\caption{Ablation study evaluating the impact of the convolution kernel shape, temporal resampling (Temp.\ res.), and the number of channels in the first network layer (Chan.), evaluated on the 5-fold cross-validation of the internal training set. Label creation is fixed to an extrusion thickness of \SI{6}{\milli\meter} and a centerline radius of \SI{4}{\milli\meter}. Values are reported as mean\,$\pm$\,standard deviation over all time points and cross-sections. Best values are in \textbf{bold}.}
\label{tab:ablation}
\scriptsize
\setlength{\tabcolsep}{4pt}
\begin{threeparttable}
\sisetup{
  table-number-alignment = center,
  separate-uncertainty   = true,
  detect-weight          = true,
  detect-inline-weight   = math
}
\begin{tabular}{@{}
    l c
    S[table-format=2.0]
    S[table-format=1.4(3)]
    S[table-format=1.3(3)]
    S[table-format=1.3(3)]
    S[table-format=1.3]
    c
    S[table-format=2.1]@{}}
\toprule
\textbf{Kernel} & \textbf{Temp.} & {\textbf{Chan.}}
& {\textbf{DSC}} & {\textbf{HD}} & {\textbf{ACD}} & {\textbf{Failed}}
& $\boldsymbol{\Delta t_{\text{train}}}$ & {$\boldsymbol{\Delta t_{\text{inf}}}$} \\
\textbf{shape} & \textbf{res.} &
& & {\si{\milli\meter}} & {\si{\milli\meter}} & {\si{\percent}}
& \textbf{fold 0} & {\si{\second}} \\
\midrule
3D cube
  & No  & 32 & 0.9157 \pm 0.064 & 2.658 \pm 1.327 & 0.806 \pm 0.582 & 0.231
  & \textbf{07:55:56} & 21.2 \\
hybrid
  & Yes & 32 & \bfseries 0.9251 \pm 0.054 & 2.435 \pm 1.205 & \bfseries 0.718 \pm 0.519 & 0.154
  & 07:59:32 & \bfseries 12.3 \\
hybrid
  & No  & 32 & 0.9236 \pm 0.061 & 2.449 \pm 1.114 & 0.722 \pm 0.499 & 0.259
  & 08:02:42 & 15.9 \\
hypercube
  & Yes & 10 & 0.9236 \pm 0.060 & 2.440 \pm 1.110 & 0.720 \pm 0.491 & 0.240
  & 08:56:17 & 15.5 \\
hypercube
  & Yes & 32 & 0.9250 \pm 0.054 & \bfseries 2.423 \pm 1.133 & 0.720 \pm 0.507 & \bfseries 0.149
  & 11:00:54 & 17.5 \\
\bottomrule
\end{tabular}
\begin{tablenotes}[flushleft]\scriptsize
\item DSC: Dice similarity coefficient; HD: Hausdorff distance;
ACD: average contour distance; $\Delta t_{\text{train}}$: training duration for
fold~0 (hh:mm:ss); $\Delta t_{\text{inf}}$: average inference time per case.
\end{tablenotes}
\end{threeparttable}
\end{table}

\begin{table}[t]
\centering
\caption{Ablation over the sparse-annotation creation parameters: extrusion thickness of the cross-section annotation and radius of the centerline, evaluated on the 5-fold cross-validation of the internal training set with the 4D hybrid kernel and temporal resampling. Values are reported as mean\,$\pm$\,standard deviation over all time points and cross-sections. Best values are in \textbf{bold}.}
\label{tab:ablation_sampling}
\scriptsize
\setlength{\tabcolsep}{4pt}
\begin{threeparttable}
\sisetup{
  table-number-alignment = center,
  separate-uncertainty   = true,
  detect-weight          = true,
  detect-inline-weight   = math
}
\begin{tabular}{@{}
    S[table-format=1.0]
    S[table-format=1.0]
    S[table-format=1.3(3)]
    S[table-format=1.3(3)]
    S[table-format=1.3(3)]
    S[table-format=1.3]@{}}
\toprule
{\textbf{Extrusion thickness}} & {\textbf{Centerline radius}}
& {\textbf{DSC}} & {\textbf{HD}} & {\textbf{ACD}} & {\textbf{Failed}} \\
{\si{\milli\meter}} & {\si{\milli\meter}}
& & {\si{\milli\meter}} & {\si{\milli\meter}} & {\si{\percent}} \\
\midrule
2 & 2 & 0.920 \pm 0.067 & 2.476 \pm 1.120 & 0.755 \pm 0.540 & 0.311 \\
2 & 4 & 0.920 \pm 0.063 & 2.542 \pm 1.229 & 0.762 \pm 0.532 & 0.270 \\
4 & 2 & 0.923 \pm 0.065 & 2.426 \pm 1.060 & 0.724 \pm 0.496 & 0.312 \\
4 & 4 & 0.923 \pm 0.063 & 2.461 \pm 1.158 & 0.727 \pm 0.500 & 0.282 \\
6 & 2 & 0.924 \pm 0.063 & \bfseries 2.413 \pm 1.099 & \bfseries 0.717 \pm 0.492 & 0.292 \\
6 & 4 & \bfseries 0.925 \pm 0.054 & 2.435 \pm 1.205 & 0.718 \pm 0.519 & \bfseries 0.154 \\
\bottomrule
\end{tabular}
\begin{tablenotes}[flushleft]\scriptsize
\item DSC: Dice similarity coefficient; HD: Hausdorff distance;
ACD: average contour distance.
\end{tablenotes}
\end{threeparttable}
\end{table}

\subsection{Comparison to Other Methods on the Test Set and the External Test Set}

\begin{table}[t]
\centering
\caption{Segmentation performance on the test set and the external test set for different volumetric aortic segmentation methods. We state whether a method produces a dynamic (4D) segmentation, which training labels were used, and which network architecture was employed. Values are reported as mean\,$\pm$\,standard deviation over all time points and cross-sections. Best values per test set are in \textbf{bold}.}
\label{tab:test_sets}
\scriptsize
\setlength{\tabcolsep}{2.5pt}
\begin{threeparttable}
\sisetup{
  table-number-alignment = center,
  separate-uncertainty   = true,
  detect-weight          = true,
  detect-inline-weight   = math
}
\begin{tabular}{@{}l l c l l
                S[table-format=1.3(3)]
                S[table-format=1.3(3)]
                S[table-format=1.3(3)]
                S[table-format=2.2]@{}}
\toprule
& \textbf{Method} & \textbf{Dyn.} & \textbf{Labels} & \textbf{Architecture}
& {\textbf{DSC}} & {\textbf{HD}} & {\textbf{ACD}} & {\textbf{Failed}} \\
& & & & & & {\si{\milli\meter}} & {\si{\milli\meter}} & {\si{\percent}} \\
\midrule
\multirow{7}{*}{\rotatebox[origin=c]{90}{\textbf{Test set}}}
& PCMRA3D    & \ding{55} & ---           & ---
  & 0.893 \pm 0.055 & 4.003 \pm 2.479 & 1.104 \pm 0.755 & \bfseries 0.00 \\
& ManReg4D   & \ding{51} & ---           & ---
  & {---} & {---} & {---} & {---} \\
& nnU3D-syn  & \ding{51} & Synth.~\cite{wolkerstorfer2026synthetically}        & nnU-Net 3D~\cite{isensee2021nnu}
  & 0.552 \pm 0.412 & 4.376 \pm 2.741 & 1.700 \pm 1.343 & 34.82 \\
& nnU3D-syn+ & \ding{51} & Synth.+~\cite{wolkerstorfer2026synthetically}       & nnU-Net 3D~\cite{isensee2021nnu}
  & 0.754 \pm 0.291 & 4.284 \pm 2.620 & 1.668 \pm 1.369 & 11.73 \\
& UTR3D-sp   & \ding{51} & Sparse (ours) & UNETR 3D~\cite{hatamizadeh2022unetr}
  & 0.885 \pm 0.149 & 3.029 \pm 1.670 & 0.926 \pm 0.663 & 2.45 \\
& nnU3D-sp   & \ding{51} & Sparse (ours) & nnU-Net 3D~\cite{isensee2021nnu}
  & 0.919 \pm 0.038 & 2.664 \pm 1.173 & 0.797 \pm 0.516 & \bfseries 0.00 \\
& nnU4D-sp   & \ding{51} & Sparse (ours) & U-Net 4D (ours)
  & \bfseries 0.927 \pm 0.033 & \bfseries 2.460 \pm 1.040 & \bfseries 0.723 \pm 0.472 & \bfseries 0.00 \\
\midrule
\multirow{7}{*}{\rotatebox[origin=c]{90}{\textbf{Ext.\ test set}}}
& PCMRA3D    & \ding{55} & ---           & ---
  & {---} & {---} & {---} & {---} \\
& ManReg4D   & \ding{51} & ---           & ---
  & 0.808 \pm 0.115 & 5.063 \pm 2.143 & 1.874 \pm 0.885 & \bfseries 0.49 \\
& nnU3D-syn  & \ding{51} & Synth.~\cite{wolkerstorfer2026synthetically}        & nnU-Net 3D~\cite{isensee2021nnu}
  & 0.499 \pm 0.402 & 6.510 \pm 5.141 & 2.481 \pm 1.943 & 37.83 \\
& nnU3D-syn+ & \ding{51} & Synth.+~\cite{wolkerstorfer2026synthetically}       & nnU-Net 3D~\cite{isensee2021nnu}
  & 0.686 \pm 0.316 & 6.124 \pm 4.188 & 2.423 \pm 1.661 & 16.06 \\
& UTR3D-sp   & \ding{51} & Sparse (ours) & UNETR 3D~\cite{hatamizadeh2022unetr}
  & 0.681 \pm 0.352 & 4.552 \pm 3.243 & 1.622 \pm 1.533 & 18.96 \\
& nnU3D-sp   & \ding{51} & Sparse (ours) & nnU-Net 3D~\cite{isensee2021nnu}
  & 0.847 \pm 0.205 & 3.437 \pm 2.147 & 1.129 \pm 1.086 & 4.41 \\
& nnU4D-sp   & \ding{51} & Sparse (ours) & U-Net 4D (ours)
  & \bfseries 0.911 \pm 0.104 & \bfseries 2.630 \pm 1.139 & \bfseries 0.757 \pm 0.596 & 0.73 \\
\bottomrule
\end{tabular}
\begin{tablenotes}[flushleft]\scriptsize
\item DSC: Dice similarity coefficient; HD: Hausdorff distance;
ACD: average contour distance; Dyn.: dynamic (time-resolved) segmentation.
``---'' indicates not applicable.
\end{tablenotes}
\end{threeparttable}
\end{table}

Table~\ref{tab:test_sets} shows the segmentation performance on both test sets. Trained on the proposed sparse labels, the 3D nnU-Net (nnU3D-sp) outperformed the publicly available 3D nnU-Net models trained on synthetic data \cite{wolkerstorfer2026synthetically} on all metrics. The proposed 4D U-Net achieved the best DSC, HD and ACD on both test sets, with a small margin over nnU3D-sp internally and a larger one externally, outperforming both manually corrected reference segmentations, i.e.\ the static PCMRA3D internally and the temporally propagated ManReg4D externally. It segmented every annotated cross-section and time frame of the internal test set and failed for 45 of 6170 contours (0.73\%) externally. Per-position results (\ref{app:perposition}) show that nnU3D-syn failed for the majority of the ascending-aorta cross-sections internally, which nnU3D-syn+ partly resolves, and that ManReg4D reached its lowest positional DSC in the distal descending aorta (0.685 vs.\ 0.879 for nnU4D-sp).

\subsection{Performance of 3D vs 4D U-Net Across Cardiac Phases}

\begin{figure}
    \centering
    \includegraphics[width=0.9\linewidth]{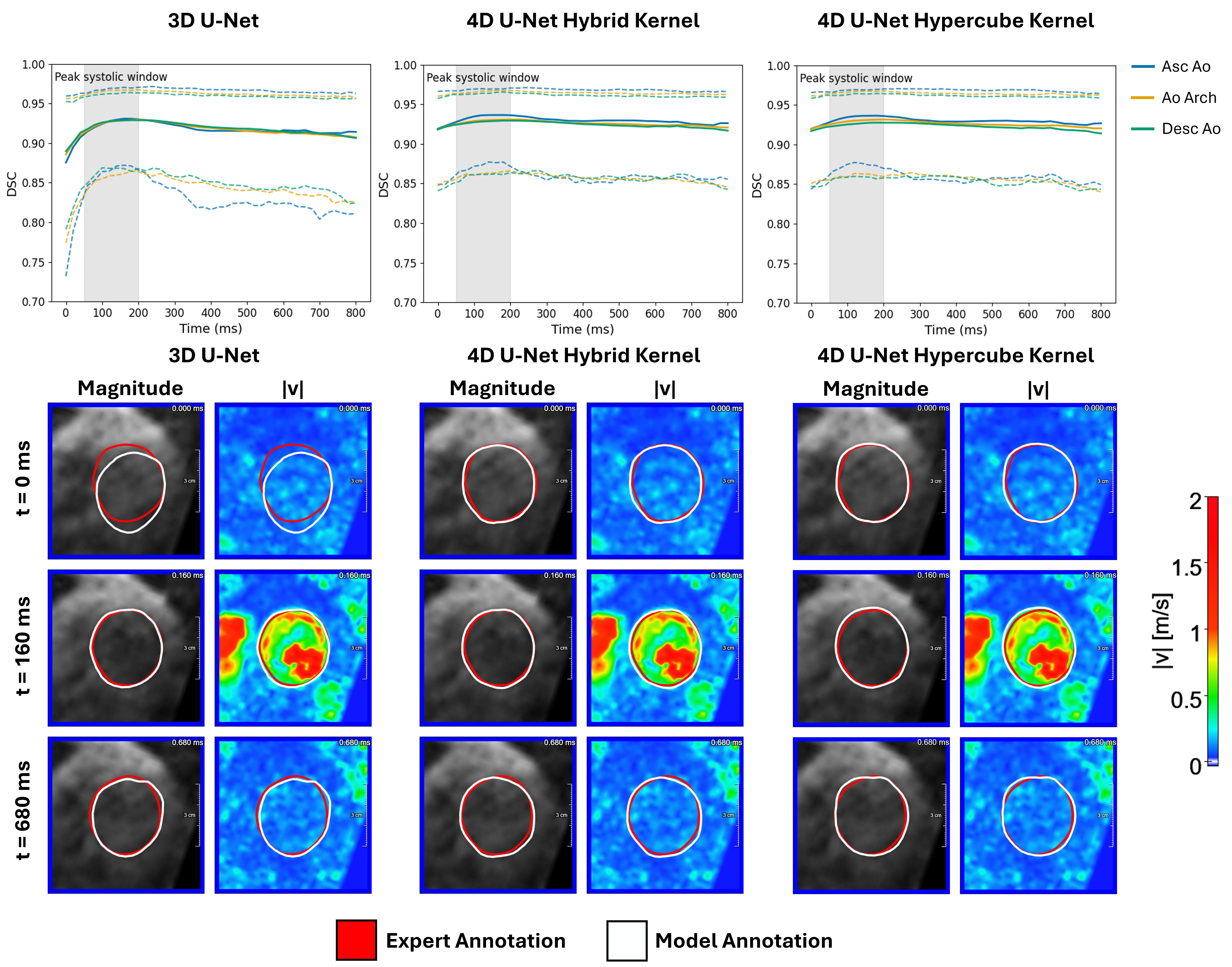}
    \caption{Comparison of 3D U-Net and 4D U-Net with hybrid kernel and hypercube kernel for different cardiac phases. The graphs show the Dice similarity coefficient (DSC) over time, evaluated using 5-fold cross-validation on the internal training set. The solid line shows the mean DSC, and the dashed lines represent the 5th and 95th percentiles. Below, the 3D U-Net and the 4D U-Net are shown for a cross-section in the A3.1 plane at end diastole ($t=0$~\si{\milli\second}), systole, and late diastole; for the time-resolved 3D segmentation and all time points see supplementary video~2.}
    \label{fig:dice_over_time}
\end{figure}

Figure~\ref{fig:dice_over_time} shows the DSC over the cardiac cycle. During systole, the 3D and 4D U-Nets performed comparably, in early cardiac phases and phases with low flow, the 4D U-Nets reached a higher DSC with no relevant difference between the hybrid kernel and the hypercube kernel. 
Performance was similar in ascending aorta, aortic arch, and descending aorta, with slightly higher DSC in the ascending aorta. On the cross-section shown below, both networks segment the lumen accurately at the systolic time point, while at $t=0$~\si{\milli\second} (end diastole) the 3D U-Net fails to maintain a proper lumen boundary and the 4D U-Nets remain aligned with the expert annotation.
The failure of the 3D U-Net is likely attributable to an increased artifact level for the first cardiac phases. These artifacts are caused by magnetization that has not yet reached steady state when the first k-space lines of the earliest cardiac phases are acquired.
The qualitative evaluation of the test set case with the lowest DSC (supplementary video~3) shows a similar behavior.
Appendix Figure~\ref{fig:dice_over_time_characteristics} stratifies the DSC over time by dataset characteristics.

\subsection{Quantitative Parameter Evaluation}
\begin{table}[htbp]
\centering
\scriptsize
\setlength{\tabcolsep}{3pt}
\caption{Agreement of the extracted quantitative parameters with the expert annotation on the test set and the external test set. Lowest absolute bias, lowest $1.96\sigma$, and highest ICC(2,1) (95\,\% CI) per parameter within each test set are highlighted in bold. ``--'' indicates that a method was not applicable to that set. Only the three methods with the lowest failure rate of the respective test set are included, and all values are computed on the cross-sections that were successfully segmented by all included methods.} \label{tab:quantitative_parameter_evaluation}
\begin{tabular}{
  ll l
  S[table-format=-2.3] S[table-format=2.3] l
  S[table-format=-2.3] S[table-format=2.3] l
}
\toprule
 & & & \multicolumn{3}{c}{Test set} & \multicolumn{3}{c}{External test set} \\
\cmidrule(lr){4-6}\cmidrule(lr){7-9}
Parameter & Unit & Method & {Bias} & {$1.96\sigma$} & {ICC (95\,\% CI)} & {Bias} & {$1.96\sigma$} & {ICC (95\,\% CI)} \\
\midrule
\multirow{4}{*}{$v_{\max}$} & \multirow{4}{*}{$\frac{\mathrm{m}}{\mathrm{s}}$}
  & nnU4D-sp & \bfseries 0.009 & \bfseries 0.130 & \bfseries 0.991 (0.99-0.99) & \bfseries 0.000 & \bfseries 0.000 & \bfseries 1.000 (1.00-1.00) \\
& & nnU3D-sp & \bfseries 0.009 & 0.133 & \bfseries 0.991 (0.99-0.99) & \bfseries 0.000 & \bfseries 0.000 & \bfseries 1.000 (1.00-1.00) \\
& & PCMRA3D  & -0.065 & 0.502 & 0.881 (0.85-0.91) & {--} & {--} & {--} \\
& & ManReg4D & {--} & {--} & {--} & 0.003 & 0.031 & 0.999 (0.998-1.00) \\
\addlinespace
\multirow{4}{*}{Net Flow} & \multirow{4}{*}{mL}
  & nnU4D-sp & 0.53 & 7.25 & 0.985 (0.98-0.99) & 0.91 & \bfseries 7.42 & \bfseries 0.990 (0.99-0.99) \\
& & nnU3D-sp & 0.36 & \bfseries 7.20 & \bfseries 0.986 (0.98-0.99) & \bfseries 0.07 & 8.01 & 0.989 (0.99-0.99) \\
& & PCMRA3D  & \bfseries 0.34 & 13.41 & 0.952 (0.94-0.96) & {--} & {--} & {--} \\
& & ManReg4D & {--} & {--} & {--} & 7.73 & 20.72 & 0.889 (0.68-0.95) \\
\addlinespace
\multirow{4}{*}{$WSS_{axial}$} & \multirow{4}{*}{Pa}
  & nnU4D-sp & \bfseries -0.008 & \bfseries 0.124 & \bfseries 0.963 (0.96-0.97) & \bfseries 0.004 & \bfseries 0.046 & \bfseries 0.985 (0.98-0.99) \\
& & nnU3D-sp & \bfseries -0.008 & 0.130 & 0.959 (0.95-0.97) & 0.005 & 0.050 & 0.983 (0.98-0.99) \\
& & PCMRA3D  & 0.051 & 0.285 & 0.713 (0.62-0.78) & {--} & {--} & {--} \\
& & ManReg4D & {--} & {--} & {--} & 0.066 & 0.122 & 0.800 (0.21-0.92) \\
\addlinespace
\multirow{4}{*}{$WSS_{circ}$} & \multirow{4}{*}{Pa}
  & nnU4D-sp & \bfseries 0.005 & 0.065 & 0.967 (0.96-0.97) & \bfseries 0.001 & \bfseries 0.026 & \bfseries 0.993 (0.99-0.99) \\
& & nnU3D-sp & 0.006 & \bfseries 0.064 & \bfseries 0.968 (0.96-0.97) & 0.002 & 0.027 & 0.992 (0.99-0.99) \\
& & PCMRA3D  & 0.014 & 0.104 & 0.901 (0.87-0.92) & {--} & {--} & {--} \\
& & ManReg4D & {--} & {--} & {--} & 0.029 & 0.075 & 0.891 (0.67-0.95) \\
\addlinespace
\multirow{4}{*}{$D(t_{Syst})$} & \multirow{4}{*}{mm}
  & nnU4D-sp & 0.43 & \bfseries 3.03 & \bfseries 0.957 (0.94-0.97) & 0.32 & \bfseries 2.99 & \bfseries 0.980 (0.97-0.98) \\
& & nnU3D-sp & \bfseries 0.29 & 3.29 & 0.952 (0.94-0.96) & \bfseries 0.08 & 4.04 & 0.964 (0.95-0.97) \\
& & PCMRA3D  & -0.62 & 4.05 & 0.932 (0.91-0.95) & {--} & {--} & {--} \\
& & ManReg4D & {--} & {--} & {--} & 2.53 & 5.01 & 0.909 (0.53-0.97) \\
\addlinespace
\multirow{4}{*}{$D(t_{EndSyst})$} & \multirow{4}{*}{mm}
  & nnU4D-sp & 0.44 & \bfseries 3.17 & \bfseries 0.954 (0.94-0.96) & \bfseries 0.14 & \bfseries 2.88 & \bfseries 0.982 (0.98-0.99) \\
& & nnU3D-sp & \bfseries 0.17 & 3.34 & 0.951 (0.94-0.96) & 0.26 & 5.45 & 0.938 (0.92-0.95) \\
& & PCMRA3D  & -0.72 & 4.18 & 0.926 (0.90-0.95) & {--} & {--} & {--} \\
& & ManReg4D & {--} & {--} & {--} & 3.61 & 4.49 & 0.869 (0.04-0.96) \\
\bottomrule
\end{tabular}
\end{table}

Table~\ref{tab:quantitative_parameter_evaluation} summarizes the agreement of the extracted quantitative parameters for the three methods with the lowest failure rate of the respective test set, computed on the cross-sections successfully segmented by all included methods ($n=366$ of 366 internally, $n=196$ of 222 externally). nnU4D-sp achieved excellent agreement for all parameters on both test sets (ICC $\geq 0.954$ internally, $\geq 0.980$ externally). Internally, the differences to nnU3D-sp were small and the 95\% confidence intervals overlapped for all parameters, whereas externally nnU4D-sp achieved highest or equal ICC and the smallest $1.96\sigma$ for all parameters. Compared with the static PCMRA3D internally and with ManReg4D externally, nnU4D-sp achieved higher ICCs for all parameters, with overlapping confidence intervals only for the aortic diameters and, externally, the maximum velocity. Although DSC, HD and ACD were better on the internal than on the external test set (Table~\ref{tab:test_sets}), the ICCs were higher externally. Bland-Altman plots for all parameters and both test sets are provided in \ref{app:bland_altman}.

\subsection{Qualitative Evaluation}

\begin{figure}
    \centering
    \includegraphics[width=1\linewidth]{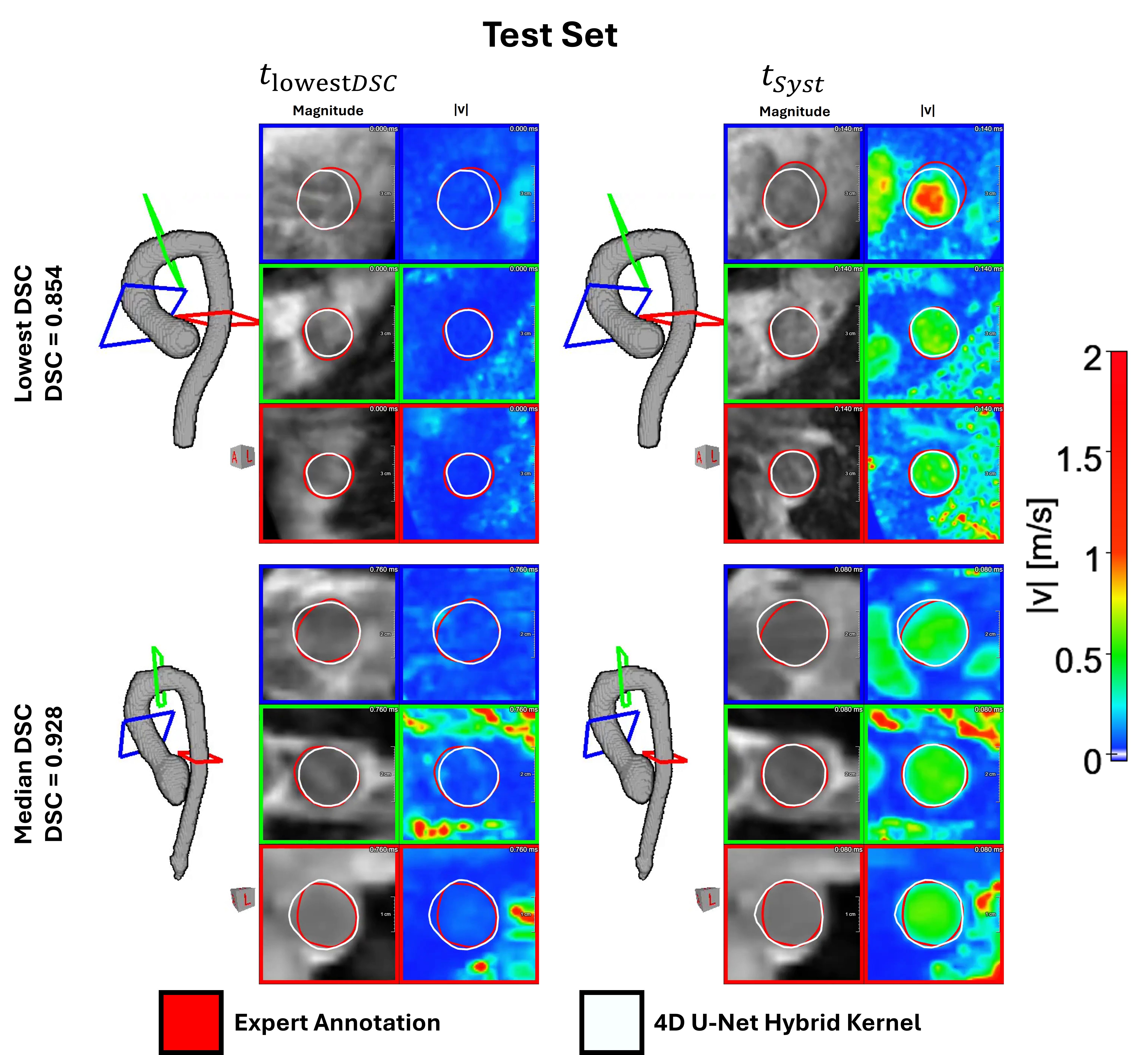}
    \caption{Qualitative evaluation on the test set. The cases with the lowest and the median DSC are shown. For each case, the 3D segmentation at the systolic time point and at the time point with the lowest DSC are displayed. The expert annotation is overlaid on the automatic segmentation in the ascending aorta, aortic arch, and descending aorta. Supplementary videos~3-5 for these cases and for the 25th DSC percentile case are provided and additionally show the results of the comparison methods.}
    \label{fig:qualitative_TS}
\end{figure}

For the qualitative evaluation we recommend the supplementary videos~3-8, as they cover all cardiac phases and the segmentations of all comparison methods.

\paragraph{Test Set}
Figure~\ref{fig:qualitative_TS} shows the cases with the lowest and the median DSC on the test set. In the lowest DSC case, nnU4D-sp underestimates the ascending aorta, at the systolic time point apparently because the network aligns the contour with the velocity magnitude rather than with the magnitude image. Supplementary video~3 shows that nnU3D-sp produces a displaced segmentation for the first two time points, while nnU3D-syn and nnU3D-syn+ fail completely for these time points. In the median DSC case the segmentations are generally well aligned, with a slight oversegmentation of the lumen in the ascending and descending aorta. Supplementary video~5 shows that nnU4D-sp moves more smoothly over the cardiac cycle than the jittering frame-wise methods, and that nnU3D-syn+ is temporally consistent but oversegments the aorta at all positions.

\paragraph{External Test Set}

\begin{figure}
    \centering
    \includegraphics[width=1\linewidth]{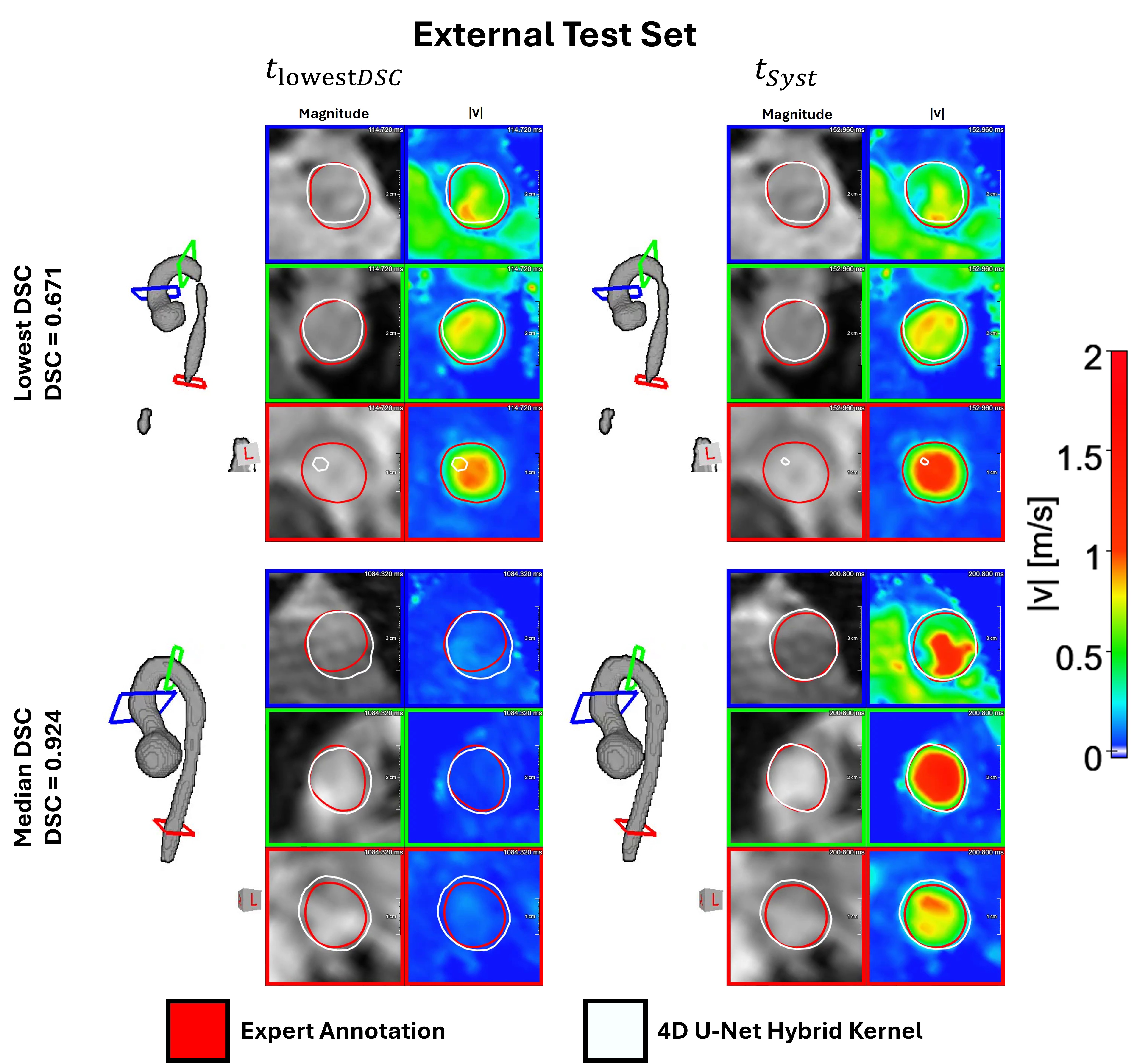}
    \caption{Qualitative evaluation on the external test set. The cases with the lowest and the median DSC are shown. For each case, the 3D segmentation at the systolic time point and at the time point with the lowest DSC are displayed. The expert annotation is overlaid on the automatic segmentation in the ascending aorta, aortic arch, and descending aorta. Supplementary videos~6-8 for these cases and for the 25th DSC percentile case are provided and additionally show the results of the comparison methods.}
    \label{fig:qualitative_ETS}
\end{figure}

Figure~\ref{fig:qualitative_ETS} shows the cases with the lowest and the median DSC on the external test set. The lowest DSC case shows two problems. The automatic ROI localization produced floating islands outside the aorta, which enlarged the bounding box (over-cropping, Section~\ref{sec:roi_cropping_evaluation}), and the segmentation of the descending aorta failed, while the ascending aorta and the arch are similar to the expert annotation. Only ManReg4D produced a valid segmentation for all time points of this case, but underestimated the lumen in the diastolic frames. In the median DSC case the segmentation agrees well with the expert annotation in the systolic frame but overestimates the lumen in the late diastolic frame. We observed this late-diastolic overestimation in several external cases, in particular in the distal descending aorta. The 25th DSC percentile case provided as a supplementary video~7 is a representative example.

\subsection{Evaluation of the ROI Cropping}
\label{sec:roi_cropping_evaluation}

Compared with processing the complete field of view, the ROI cropping improved segmentation accuracy on both test sets (DSC 0.927 vs.\ 0.921 internally, 0.911 vs.\ 0.894 externally) and reduced the end-to-end runtime by a factor of 2.2 internally and 2.4 externally. No under-cropping occurred, i.e.\ all sparse reference segmentations were fully contained in the cropped region. Floating islands of the localizer enlarged the bounding box (over-cropping) in 36.67\% of the external test set cases (Table~\ref{tab:cropping_evaluation}, \ref{app:localizer}).

\section{Discussion}
\label{sec:discussion}

We present a fully automated 4D CNN for time-resolved aortic segmentation in 4D flow MRI, trained exclusively on sparse labels derived from time-resolved 2D cross-sectional contours and static centerlines. On the internal test set it reached a DSC of $0.927 \pm 0.033$ without a single failed cross-section, and it generalized to an external post-contrast cohort from another site with a different protocol and a different annotator (DSC $0.911 \pm 0.104$, 0.73\,\% failures), with excellent agreement of all derived hemodynamic parameters (ICC $\geq 0.954$ internally, $\geq 0.980$ externally).

The 4D network outperformed the frame-wise 3D baselines on all segmentation metrics, with a larger gain externally than internally and especially a lower failure rate externally (0.73\% vs. 4.41\%). This suggests that temporal context can act as a regularizer under domain shift. The phase-resolved analysis shows that this improvement originates from low-flow phases, where vessel–background contrast is weak and temporal context compensates for missing spatial evidence. For the extracted parameters, differences were small and ICC confidence intervals overlapped for most parameters. This is plausible, as the established parameters are dominated by high-flow time points, exactly the phases in which the frame-wise network already performs well. The benefit of the 4D convolutions therefore lies in three other aspects: robustness to failure, a reduction of inference time by a factor of 1.7, enabled by compressing the temporal dimension in the network's latent space, and a temporally more consistent segmentation in the diastolic frames. The latter is a prerequisite for analyses that integrate or differentiate over the full cardiac cycle, such as pathline-based flow-component analyses, moving-boundary computational fluid dynamics, and statistical shape/motion models. Demonstrating the impact on these downstream applications was beyond the scope of this work and remains for future studies.

4D hypercube and hybrid kernels performed equivalently during cross-validation, so the benefit of the hybrid kernel is efficiency rather than accuracy. It uses 29 instead of 81 weights per channel combination, leading to about 27\,\% shorter training and 30\,\% shorter inference time, especially relevant for hardware weaker than the H200 GPUs used here. Matching the parameter count by reducing the hypercube channels instead reduced accuracy. Replacing the five-fold ensemble by a single model and removing test-time augmentation is a further straightforward reduction for deployment, but may reduce segmentation performance.

Compared with semi-automatic methods the network outperformed a static PCMRA segmentation on the internal test set and a temporally registered segmentation method on the external test set. It achieved higher DSC and larger ICCs for all extracted hemodynamic parameters. For the static PCMRA segmentation it must be noted that it was created for centerline and plane definition and not for extracting segmentation-sensitive parameters, so its lower agreement is no upper bound for a carefully created static segmentation.

Sparse-label training clearly outperformed synthetic training on both test sets, even when using the same 3D architecture. The gap to the in-domain performance of the synthetically trained models is most plausibly explained by domain shift, differing pathologies, a different reconstruction (locally low-rank / FlowMRI-Net \cite{wolkerstorfer2026synthetically}), and a considerably older target cohort, i.e.\ by the diversity of the training set rather than by label sparsity itself. The 268 scans from eight centers, two vendors, seven scanner models, several protocols, heterogeneous valve morphologies, and post-surgical states were only attainable because the labels are sparse and already existed from cross-sectional analyses. Sparse annotation should therefore be seen as an enabler of heterogeneous multicenter data rather than as a compromise. The 25th-percentile external case (supplementary video~7) illustrates both benefit and limits: our model produced an acceptable segmentation of an aortic isthmus stenosis for which both synthetically trained models fail, but reproduced a near-circular cross-section at the stenosis, where the true lumen is not circular. The network has evidently learned a strong shape prior and enforces it, which stabilizes low-contrast diastolic frames but systematically biases non-circular pathological geometries.

It is initially surprising that the largest extrusion thickness ($et = 6$\,mm) performed best, although sub-voxel deviations can change WSS by more than 20\,\% (Figure~\ref{fig:introduction}). The annotation planes are placed perpendicular to the centerline. Extrusion along the plane normal is a good approximation wherever the centerline is locally straight, the diameter changes slowly and the plane orientation is accurate. So thicker extrusion mainly increases the number of supervised voxels, consistent with the lower failure rate. When transferring the scheme to other vessels or datasets, the extrusion thickness should be treated as a dataset-specific hyperparameter, with the search range adapted to voxel size, vessel curvature and the angular accuracy of the planes.

\subsection{Limitations}
\label{subsec:discussion_limitations}

First, most nnU-Net design choices (preprocessing, spatial resampling target, normalization, augmentation, loss, optimizer, sampling, inference) were adopted without dedicated ablation. With a single dataset we can show that they transfer well to a 4D network, not that they are generally optimal.
Second, evaluation is based on 2D+time cross-sections. DSC agreed well with its volumetric counterpart, whereas the 2D+time HD is a relative measure only, and the excellent parameter ICCs must be read cautiously because they are dominated by high-flow phases.
Third, the learned near-circular shape prior can distort stenoses, dissections, coarctations and aneurysm necks, and our cohorts contain too few such geometries to quantify this bias, so targeted evaluation is required before shape-sensitive quantification in these patients.
Fourth, supervision covered only the ascending aorta, arch and proximal descending aorta and only sagittal and oblique sagittal acquisitions (tilt toward coronal $21.8 \pm 8.0^\circ$ internally and $22.7 \pm 10.1^\circ$ externally). Accordingly, late-diastolic lumen overestimation occurred in the distal descending aorta externally, supra-aortic branches are not represented, and fundamentally different orientations such as axial whole-heart coverage require separate validation, whereas axis flips from differing coordinate conventions should not affect the model because mirroring-based test-time augmentation is applied along all spatial axes \cite{isensee2021nnu}.
Finally, expert oversight remains necessary: in the worst external case the descending aorta segmentation failed over parts of the cycle, an error detectable by visual inspection or automatic topology checks.

\section{Conclusion}
We developed and evaluated a fully automated method for 3D+time aortic segmentation in 4D flow MRI that addresses two key challenges: the joint modeling of three spatial dimensions and time, and the scarcity of dense 4D training annotations. A hybrid 4D convolution captures temporal dependencies at a low parameter and runtime cost, and sparse 4D labels derived from expert-annotated 2D+time cross-sections enable training on a diverse multicenter cohort. The model generalized to an external test set acquired post-contrast with a different MRI protocol and achieved excellent agreement for wall shear stress, net flow, and aortic diameters. To promote reproducibility, we publicly release the trained model. The approach is readily extensible to other vascular regions, such as the carotid arteries, and is a step toward scalable, automated 4D flow MRI analysis.

\section*{Abbreviations}
2D, two-dimensional; 3D, three-dimensional; 4D, four-dimensional; ACD, average contour distance; ASD, average surface distance; BAV, bicuspid aortic valve; bSSFP, balanced steady-state free precession; CI, confidence interval; CNN, convolutional neural network; cr, centerline radius; DSC, Dice similarity coefficient; et, extrusion thickness; FoV, field of view; HD, Hausdorff distance; HD95, 95th-percentile Hausdorff distance; ICC, intraclass correlation coefficient; LoA, limits of agreement; MPR, multiplanar reconstruction; MRI, magnetic resonance imaging; PC, phase contrast; PC-MRA, phase-contrast magnetic resonance angiography; PC-MRI, phase-contrast magnetic resonance imaging; ROI, region of interest; VENC, velocity encoding; WSS, wall shear stress

\section*{Data availability}
The software code and instructions to run inference with the complete inference pipeline and trained models can be found under \url{https://github.com/hinrah/Aorta_Segmentation_in_4D_flow_MRI.git}. Trained models will be available upon manuscript acceptance.
The software code for the 4D U-Net, as well as all software used for training is available at \url{https://github.com/hinrah/nnunet4D.git}.
The data used for training and evaluation is not publicly available due to privacy.

\section*{Acknowledgement}
The authors acknowledge the Scientific Computing of the IT Division at the Charité - Universitätsmedizin Berlin for providing computational resources that have contributed to the research results reported in this paper.  \url{https://www.charite.de/en/research/research_support_services/research_infrastructure/science_it/#c30646061}

\section*{Compliance with Ethical Standards}

\paragraph*{Funding}
This work was funded by the German Research Foundation (GRK2260, BIOQIC). H.R. and J.S.-M. received support from the DZHK project 81Z0100229. S.K. received support from the DZHK (German Center for Cardiovascular Research), Partner Site Berlin. S.K. was supported by an unrestricted research grant from Philips Healthcare. T.K., S.K., and A.He. were partially funded by the German Research Foundation - CRC-1470 - 437531118. A.He. was partially funded by the German Research Foundation grant \#HE7312/7-1 515294457. A.Ha. was supported by the Berta-Ottenstein-Program for Advanced Clinician Scientists, Faculty of Medicine, University of Freiburg, Germany and by the German Research Foundation \#HA 5399/6-1.

\paragraph{Conflict of Interest:}
Sebastian Kelle reports a relationship with Philips Healthcare that includes funding grants. The other authors declare that they have no known competing financial interests or personal relationships that could have appeared to influence the work reported in this paper.

\paragraph{Informed Consent:}
The studies involving human participants were reviewed and approved by the local ethics boards of Albert-Ludwigs-Universität Freiburg, Charité – Universitätsmedizin Berlin, Technical University of Munich, University Medical Center Hamburg-Eppendorf, and University of Calgary. All datasets analyzed were acquired prospectively in the context of these approved studies, and written informed consent was obtained from all participants prior to data acquisition. The original ethics approvals and consent forms cover secondary analysis of the data for research purposes.

\section*{Declaration of generative AI and AI-assisted technologies in the manuscript preparation process}
\paragraph{Statement} During the preparation of this work the authors used Claude Opus5 to assist with table and formular formatting. After using this service, the authors reviewed and edited the content as needed and take full responsibility for the content of the published article.

\bibliographystyle{elsarticle-num}
\bibliography{references.bib}
\newpage

\appendix
\section{Localization Network for ROI Cropping}
\label{app:localizer}

The 3D localization network used for the ROI cropping (Section~\ref{sec:cropping}) solves a static binary segmentation task: given the time-averaged PC-MRA of a 4D flow acquisition, it predicts a single 3D aorta mask. This mask is used exclusively to derive the crop.

\subsection{ROI Cropping Methodology}
\paragraph{Data and PC-MRA computation}
The localizer was trained on the same cases as the 4D segmentation network, i.e.\ the internal training set described in Section~\ref{sec:data} ($N=\num{268}$ scans), using the manual 3D PC-MRA annotations of Section~\ref{sec:annotation} as labels. The PC-MRA is computed from the magnitude images $M(t)$ and the velocity magnitude $|\mathbf{v}(t)|=\sqrt{v_x^2+v_y^2+v_z^2}$ as $\mathrm{PCMRA}=\frac{1}{T}\sum_{t=1}^{T}{M}(t)^{2}\,|\mathbf{v}(t)|^{2}$, averaged over all $T$ reconstructed cardiac phases.

\paragraph{Architecture and training}
We used the default nnU-Net~\cite{isensee2021nnu} 3d\_fullres configuration (nnU-Net v2, nnUNetTrainer, nnUNetPlans) as generated by the automatic experiment planner. Images are resampled to the target spacing of $\num{2.5}\times\num{2.125}\times\num{2.125}\,\si{\milli\metre\cubed}$ and z-score normalized.
The network is a plain convolutional U-Net with \num{6} stages ($\num{32}\to\num{64}\to\num{128}\to\num{256}\to\num{320}\to\num{320}$ features), two 3D cube convolutions per stage, instance normalization, leaky ReLU and deep supervision, with isotropic downsampling in stages~2-4 and in-plane downsampling in stages~5-6. The patch size is $\num{32}\times\num{160}\times\num{128}$ voxels (batch size \num{8}), which covers the median training volume of $\num{30}\times\num{158}\times\num{118}$ voxels. Training followed the nnU-Net defaults without modification (\num{1000} epochs of \num{250} mini-batches, Dice\,+\,cross-entropy loss, stochastic gradient descent with Nesterov momentum 0.99, initial learning rate 0.01, nnU-Net learning-rate decay, default augmentation) in a subject-wise 5-fold cross-validation on NVIDIA H200 GPUs. Note that the localizer, unlike the 4D models, was trained for the full nnU-Net default of 1000 epochs, as only a single 3D configuration had to be trained.

\paragraph{Inference and derivation of the ROI}
At test time the five fold models are applied as an ensemble by averaging their softmax outputs, using sliding-window inference with a tile step size of \num{0.5}, Gaussian tile weighting, and mirroring along all spatial axes \cite{isensee2021nnu}. No postprocessing is applied; in particular, connected-component filtering was deliberately not enabled to prevent potential under-cropping. The ROI is the bounding box of all predicted foreground voxels, extended by \num{5} voxels and clipped at the image boundaries. All time frames are cropped with the identical box. After 4D inference on the cropped ROI, the predicted 4D labels are transformed back to the original image geometry, so that all reported metrics are computed in the original image space. The same localizer, trained on the internal training set, was applied unchanged to both test sets and was never fine-tuned or adapted.

\paragraph{Cross-validation results}
Table~\ref{tab:localizer} reports the cross-validation performance of the localizer against the manual PC-MRA segmentations. The values quantify the accuracy of the static 3D mask. The operative requirement on the localizer is that the derived bounding box contains the aorta, which is evaluated on both test sets in the main text.

\subsection{Detailed evaluation of the ROI Cropping on the test sets}

\begin{table}[t]
\centering
\caption{Post-hoc evaluation of the ROI cropping (bounding box of the 3D time-averaged segmentation with a \num{5}-voxel margin) on the test set and the external test set. The cropping was fixed before any test-set evaluation. The comparison against processing the full field of view was performed post-hoc and was not used for any model or parameter selection. Runtimes are given per case. Best values per test set are in \textbf{bold}.} \label{tab:cropping_evaluation}
\scriptsize
\setlength{\tabcolsep}{3pt}
\begin{threeparttable}
\sisetup{
  table-number-alignment = center,
  detect-weight          = true,
  detect-inline-weight   = math
}
\begin{tabular}{@{}
    l c
    S[table-format=1.3]
    S[table-format=1.3]
    S[table-format=1.3]
    S[table-format=1.2]
    S[table-format=2.2]
    S[table-format=3.2]
    S[table-format=3.2]
    S[table-format=2.2]
    S[table-format=2.2]@{}}
\toprule
& & \multicolumn{4}{c}{\textbf{Performance}}
& \multicolumn{3}{c}{\textbf{Runtime} (\si{\second}\,/\,case)}
& \multicolumn{2}{c}{\textbf{Localization} (\si{\percent})} \\
\cmidrule(lr){3-6}\cmidrule(lr){7-9}\cmidrule(l){10-11}
{} & {\textbf{Crop}}
& {\textbf{DSC}} & {\textbf{HD}} & {\textbf{ACD}} & {\textbf{Failed}}
& {\textbf{ROI}} & {\textbf{4D}} & {\textbf{Total}}
& {\textbf{Under-}} & {\textbf{Over-}} \\
& & & {\si{\milli\meter}} & {\si{\milli\meter}} & {\si{\percent}}
& {\textbf{steps}} & {\textbf{seg.}} & & {\textbf{crop}} & {\textbf{crop}} \\
\midrule
\multirow{2}{*}{Test set}
  & \ding{55}   & 0.921 & 2.623 & 0.793 & 0.00 & {--} & 40.81 & 40.81 & {--} & {--} \\
  & \ding{51} & \bfseries 0.927 & \bfseries 2.460 & \bfseries 0.723 & 0.00 & 6.09 & \bfseries 12.29 & \bfseries 18.38 & 0.00 & 0.00 \\
\midrule
\multirow{2}{*}{Ext.\ test set}
  & \ding{55}   & 0.894 & 3.175 & 0.955 & 1.25 & {--} & 113.37 & 113.37 & {--} & {--} \\
  & \ding{51} & \bfseries 0.911 & \bfseries 2.630 & \bfseries 0.757 & \bfseries 0.73 & 8.44 & \bfseries 38.55 & \bfseries 46.99 & 0.00 & 36.67 \\
\bottomrule
\end{tabular}
\begin{tablenotes}[flushleft]\scriptsize
\item \ding{51}: with ROI cropping; \ding{55}: full field of view.
ROI steps: PC-MRA computation, 3D time-averaged segmentation, cropping and back-transformation into the original image space; 4D seg.: inference of the 4D U-Net ensemble.
\item Under-crop: fraction of cases in which at least one sparse 2D\,+\,$t$ reference segmentation is not fully contained in the cropped region.
Over-crop: fraction of cases whose bounding box is enlarged by floating islands of the 3D time-averaged segmentation.
\end{tablenotes}
\end{threeparttable}
\end{table}

\paragraph{Segmentation accuracy and runtime}
Processing the complete image instead of the cropped ROI degraded segmentation accuracy on both test sets and increased the runtime (Table~\ref{tab:cropping_evaluation}). On the test set, the DSC decreased from \num{0.927} to \num{0.921}; on the external test set the DSC decreased from \num{0.911} to \num{0.894} and the fraction of cross-sections without a valid segmentation rose from \SI{0.73}{\percent} to \SI{1.25}{\percent}. Without cropping, the 4D segmentation network had to process the full field of view, which increased its runtime from \SI{12.29}{\second} to \SI{40.81}{\second} per case on the test set and from \SI{38.55}{\second} to \SI{113.37}{\second} per case on the external test set. The additional steps introduced by the cropping (PC-MRA computation, 3D time-averaged segmentation, cropping and back-transformation into the original image space) accounted for \SI{6.09}{\second} and \SI{8.44}{\second} per case, respectively. In total, the ROI cropping reduced the end-to-end runtime by a factor of \num{2.2} on the test set (\SI{40.81}{\second} to \SI{18.38}{\second}) and by a factor of \num{2.4} on the external test set (\SI{113.37}{\second} to \SI{46.99}{\second}).

\paragraph{Localization failures} We additionally evaluated whether the ROI cropping excluded annotated vessel regions. No under-cropping occurred on either test set, i.e.\ all sparse 2D\,+\,$t$ reference segmentations were fully contained in the cropped region for every case. Over-cropping, i.e.\ a bounding box enlarged by floating islands in the 3D time-averaged segmentation, did not occur on the test set but was observed in \SI{36.67}{\percent} of the external test set cases. Over-cropping enlarges the processed volume and therefore the runtime, but the segmentation metrics DSC, HD, and ACD improved despite over-cropping in more than one third of the cases.

\begin{table}[t]
\centering
\caption{Performance of the 3D localization nnU-Net on the time-averaged PC-MRA, evaluated against the manual PC-MRA segmentations in the subject-wise five-fold cross-validation of the internal training set (all folds combined, $n=268$ scans). Values are mean $\pm$ standard deviation over scans and are computed in the original image geometry. DSC: Dice similarity coefficient; HD95: 95th-percentile Hausdorff distance; ASD: average surface distance.}
\label{tab:localizer}
\begin{tabular}{ccc}
\toprule
DSC & HD95 [mm] & ASD [mm] \\
\midrule
$0.904 \pm 0.069$ & $7.56 \pm 8.48$ & $1.45 \pm 1.34$ \\
\bottomrule
\end{tabular}
\end{table}

\section{Detailed Description of 4D Convolution Kernels}
\label{app:detailed_4d_conv}
This section uses a different notation of convolution than the main text, to be as close to the implemented version as possible. We implemented 4D convolutions by summing multiple 3D convolutions, as proposed by Myronenko et al.\ \cite{myronenko20194d}. Let $x$ denote the input feature map after padding with $\lfloor K_i/2 \rfloor$ elements on both sides of each dimension $i \in \{T,D,H,W\}$, so that the output has the same extent as the unpadded input for unit stride. Spatial padding is zero padding; temporal padding is cyclic (see below). With the strides $s_t, s_d, s_h, s_w$, which are equal to one in all layers except the strided convolutions used for downsampling, the 4D convolution is defined as

\begin{multline}
y[c_{\text{out}},\,t,\, d,\, h,\, w]
= b[c_{\text{out}}] \\
+ \sum_{c_{\text{in}}=0}^{C_{\text{in}}-1}
  \sum_{k_t=0}^{K_T-1}
  \sum_{k_d=0}^{K_D-1}
  \sum_{k_h=0}^{K_H-1}
  \sum_{k_w=0}^{K_W-1}
  W\bigl[c_{\text{out}},\, c_{\text{in}},\, k_t,\, k_d,\, k_h,\, k_w\bigr] \\
  \cdot\, x\bigl[c_{\text{in}},\;
         t \cdot s_t + k_t,\;
         d \cdot s_d + k_d,\;
         h \cdot s_h + k_h,\;
         w \cdot s_w + k_w\bigr]
\end{multline}

For $K_T = 3$ this can be reformulated as the summation over three 3D convolutions with the 3D kernel weights $W_{k_t=0}$, $W_{k_t=1}$, and $W_{k_t=2}$:

\begin{multline}
y[c_{\text{out}},\,t,\, d,\, h,\, w]
= b[c_{\text{out}}] \\
+ \sum_{c_{\text{in}}=0}^{C_{\text{in}}-1}
  \sum_{k_d=0}^{K_D-1}
  \sum_{k_h=0}^{K_H-1}
  \sum_{k_w=0}^{K_W-1}
  W_{k_t=0}\bigl[c_{\text{out}},\, c_{\text{in}},\, k_d,\, k_h,\, k_w\bigr] \\
  \cdot\, x\bigl[c_{\text{in}},\;
         t \cdot s_t + 0,\;
         d \cdot s_d + k_d,\;
         h \cdot s_h + k_h,\;
         w \cdot s_w + k_w\bigr]\\
+ \sum_{c_{\text{in}}=0}^{C_{\text{in}}-1}
  \sum_{k_d=0}^{K_D-1}
  \sum_{k_h=0}^{K_H-1}
  \sum_{k_w=0}^{K_W-1}
  W_{k_t=1}\bigl[c_{\text{out}},\, c_{\text{in}},\, k_d,\, k_h,\, k_w\bigr] \\
  \cdot\, x\bigl[c_{\text{in}},\;
         t \cdot s_t + 1,\;
         d \cdot s_d + k_d,\;
         h \cdot s_h + k_h,\;
         w \cdot s_w + k_w\bigr]\\
+ \sum_{c_{\text{in}}=0}^{C_{\text{in}}-1}
  \sum_{k_d=0}^{K_D-1}
  \sum_{k_h=0}^{K_H-1}
  \sum_{k_w=0}^{K_W-1}
  W_{k_t=2}\bigl[c_{\text{out}},\, c_{\text{in}},\, k_d,\, k_h,\, k_w\bigr] \\
  \cdot\, x\bigl[c_{\text{in}},\;
         t \cdot s_t + 2,\;
         d \cdot s_d + k_d,\;
         h \cdot s_h + k_h,\;
         w \cdot s_w + k_w\bigr]
\end{multline}

This allows the use of optimized CUDA implementations of the 3D convolutions.

Incorporating the fourth dimension increases the number of trainable weights per combination of input and output channel from 27 ($3\times3\times3$) to 81 ($3\times3\times3\times3$). Following an approach that has proven successful for sparse convolutions \cite{choy20194d}, we also implemented a hybrid kernel. It applies a full 3D cube ($3\times3\times3$) spatial kernel to the current time frame and a $1\times1\times1$ kernel to the two adjacent time frames and therefore uses only two additional trainable weights $w_{k_t=0}$ and $w_{k_t=2}$ per channel combination compared to a 3D convolution. For $K_D = K_H = K_W = 3$ the spatially central index is $\lfloor K/2 \rfloor = 1$, so the hybrid convolution is given by

\begin{multline}
y[c_{\text{out}},\,t,\, d,\, h,\, w]
= b[c_{\text{out}}] \\
+ \sum_{c_{\text{in}}=0}^{C_{\text{in}}-1}
  w_{k_t=0}\bigl[c_{\text{out}},\, c_{\text{in}}\bigr] \\
  \cdot\, x\bigl[c_{\text{in}},\;
         t \cdot s_t + 0,\;
         d \cdot s_d + 1,\;
         h \cdot s_h + 1,\;
         w \cdot s_w + 1\bigr]\\
+ \sum_{c_{\text{in}}=0}^{C_{\text{in}}-1}
  \sum_{k_d=0}^{K_D-1}
  \sum_{k_h=0}^{K_H-1}
  \sum_{k_w=0}^{K_W-1}
  W_{k_t=1}\bigl[c_{\text{out}},\, c_{\text{in}},\, k_d,\, k_h,\, k_w\bigr] \\
  \cdot\, x\bigl[c_{\text{in}},\;
         t \cdot s_t + 1,\;
         d \cdot s_d + k_d,\;
         h \cdot s_h + k_h,\;
         w \cdot s_w + k_w\bigr]\\
+ \sum_{c_{\text{in}}=0}^{C_{\text{in}}-1}
  w_{k_t=2}\bigl[c_{\text{out}},\, c_{\text{in}}\bigr] \\
  \cdot\, x\bigl[c_{\text{in}},\;
         t \cdot s_t + 2,\;
         d \cdot s_d + 1,\;
         h \cdot s_h + 1,\;
         w \cdot s_w + 1\bigr]
\end{multline}

This results in 29 ($1 + 3\times3\times3 + 1$) trainable parameters per combination of input and output channel.

\section{Detailed Description of the Sparse 4D Label Creation}
\label{app:sparse_labels}

This appendix details the generation of the sparse 4D training labels summarized in Section~\ref{sec:sparse_labels}. The formalism follows Rahlfs et al.~\cite{rahlfs5564923learning} and is adapted to (i) the creation of 3D+time instead of 3D sparse labels, (ii) the 2D+time nature of the cross-sectional annotations, (iii) a static, non-branching centerline annotation, and (iv) a two-class problem (lumen and background, no vessel wall).

\subsection{Notation}
The 4D images and the sparse training labels are defined on a discrete voxel grid
\begin{equation}
    \mathbb{D}_{4} = \mathbb{D} \times \mathbb{T}, \qquad
    \mathbb{D} \subset \mathbb{Z}^{3}, \qquad
    \mathbb{T} = \{1, 2, \ldots, T\},
\end{equation}
where $\mathbb{D}$ is the spatial grid and $\mathbb{T}$ the set of cardiac time frames. The spatial grid geometry is identical for all time frames, i.e.\ a single affine matrix $A$ (voxel-to-world matrix) maps the center point of a spatial voxel $\mathbf{v} \in \mathbb{D}$ to its world coordinates $\mathbf{w}_{\mathbf{v}} \in \mathbb{R}^{3}$. All distances are computed in world coordinates, which makes the label creation independent of anisotropic voxel spacings.

\paragraph{Cross-sectional 2D+time annotations}
The annotations consist of $M$ cross-sectional plane annotations $\Pi_m^{(\tau)}$. Each plane is defined by a normal vector $\mathbf{n}_m$ and its minimal distance $d_m$ to the origin. Since the planes were placed once on the temporally static centerline geometry while the lumen contour was annotated in every time frame, the plane parameters are constant over time and only the contour is time-dependent:
\begin{equation}
    \Pi_m^{(\tau)} = \bigl(\mathbf{n}_m,\, d_m,\, \gamma_m^{(\tau)}\bigr),
    \qquad m = 1,\ldots,M, \quad \tau \in \mathbb{T}.
\end{equation}
In contrast to \cite{rahlfs5564923learning}, only the inner (luminal) contour is annotated, i.e.\ $\gamma_m^{(\tau)} \equiv \gamma_{m,\text{inner}}^{(\tau)}$ and no outer contour exists. Each contour is a closed polygon formed by a sequence of contour points
\begin{equation}
    \gamma_m^{(\tau)} = \bigl\{ \mathbf{p}_1^{(\tau)}, \mathbf{p}_2^{(\tau)}, \ldots,
    \mathbf{p}_{n_m}^{(\tau)} \bigr\}, \qquad \mathbf{p}_i^{(\tau)} \in \mathbb{R}^{3},
\end{equation}
whose points all lie in the corresponding plane,
\begin{equation}
    \forall\, \mathbf{p}_i^{(\tau)} \in \gamma_m^{(\tau)}: \quad
    \mathbf{n}_m \cdot \mathbf{p}_i^{(\tau)} - d_m = 0 .
\end{equation}

\paragraph{Centerline annotation}
The centerline is annotated as a graph $\mathcal{G} = (\mathcal{V}, \mathcal{E})$, where each edge $e = (v_i, v_j) \in \mathcal{E}$ is associated with a polyline of skeleton points
\begin{equation}
    \mathcal{S}_{ij} = \bigl\{ \mathbf{s}_1^{(ij)}, \ldots, \mathbf{s}_{k_{ij}}^{(ij)} \bigr\},
    \qquad \mathbf{s}_k^{(ij)} \in \mathbb{R}^{3},
    \qquad \mathcal{S} = \bigcup_{i,j} \mathcal{S}_{ij}.
\end{equation}
For the aorta, the centerline is a single, non-branching path from the aortic root to the distal end of the annotated segment, i.e. $|\mathcal{E}| = 1$ and $\mathcal{S} = \mathcal{S}_e$. The centerline was annotated once on the time-averaged PC-MRA and is therefore constant over $\mathbb{T}$.

\subsection{Per-Frame Sparse Label Maps}
For every time frame $\tau$ a sparse 3D label map
\begin{equation}
    Y^{(\tau)} : \mathbb{D} \rightarrow C \cup \{\xi\}, \qquad C = \{\text{Background}, \text{Lumen}\},
\end{equation}
is created, where $\xi$ denotes the ignore label. Voxels labeled with $\xi$ do not contribute to the training loss~\cite{gotkowski2025revisiting, rahlfs5564923learning}. $Y^{(\tau)}$ is obtained by joining several intermediate sparse label maps, each derived from a subset of the annotations.

\subsubsection{Centerline-Based Labels}
The distance of a voxel $\mathbf{v}$ to the centerline is
\begin{equation}
    d_{\mathcal{S}}(\mathbf{v}) = \min_{\mathbf{s} \in \mathcal{S}} \lVert \mathbf{w}_{\mathbf{v}} - \mathbf{s} \rVert ,
\end{equation}
where the minimum is taken over a densely resampled polyline. With the centerline radius $cr$ and the background radius $br$, the intermediate label map is
\begin{equation}
Y_{\mathcal{S}}(\mathbf{v}) =
\begin{cases}
\text{Lumen}, & \text{if } d_{\mathcal{S}}(\mathbf{v}) \leq cr, \\
\text{Background}, & \text{if } d_{\mathcal{S}}(\mathbf{v}) \geq br, \\
\xi, & \text{otherwise}.
\end{cases}
\end{equation}
Because the centerline annotation is static, $Y_{\mathcal{S}}$ is identical for all time frames. This is valid as long as $cr$ is smaller than the minimum centerline-to-lumen-boundary distance and $br$ is larger than the maximum centerline-to-lumen-boundary distance over the complete cardiac cycle. Both bounds can only be estimated from the time-resolved contour annotations. Accordingly, $br$ was fixed to the largest contour-point-to-centerline distance observed in the training set (\SI{34.9}{\milli\meter}) plus a safety margin of \SI{20}{\percent}, i.e. \SI{41.9}{\milli\meter}, whereas $cr$ was treated as a free parameter and selected by ablation (\ref{app:coord}).

\subsubsection{Cross-Section-Based Labels}
The distance ($d_{\Pi_m}(\mathbf{v})$) of a voxel $\mathbf{v}$ to the plane of annotation $m$ and its orthogonal projection ($P_m(\mathbf{v})$) onto that plane are
\begin{align}
    d_{\Pi_m}(\mathbf{v}) &= \lvert \mathbf{n}_m \cdot \mathbf{w}_{\mathbf{v}} - d_m \rvert, \\
    P_m(\mathbf{v}) &= \mathbf{w}_{\mathbf{v}} - \mathbf{n}_m \bigl( \mathbf{n}_m \cdot \mathbf{w}_{\mathbf{v}} - d_m \bigr).
\end{align}
The predicate $\text{inside}(\mathbf{p}, \gamma)$ is true if $\mathbf{p}$ lies strictly inside the closed polygon $\gamma$ and false if it lies on or outside the polygon. It is evaluated in the 2D coordinate system of the plane. With the extrusion thickness parameter $et$, the intermediate label map of annotation $m$ at time frame $\tau$ is given by the wall-free variant of \cite{rahlfs5564923learning}:
\begin{equation}
Y_{\Pi_m}^{(\tau)}(\mathbf{v}) =
\begin{cases}
\text{Lumen}, & \text{if } d_{\Pi_m}(\mathbf{v}) \leq \tfrac{et}{2}
 \text{ and } \text{inside}\bigl(P_m(\mathbf{v}), \gamma_m^{(\tau)}\bigr), \\[2pt]
\text{Background}, & \text{if } d_{\Pi_m}(\mathbf{v}) \leq \tfrac{et}{2}
 \text{ and } d_{\mathcal{S} \cap \Pi_m}(\mathbf{v}) \leq br \\
 & \text{and not } \text{inside}\bigl(P_m(\mathbf{v}), \gamma_m^{(\tau)}\bigr), \\[2pt]
\xi, & \text{otherwise}.
\end{cases}
\end{equation}
Note that $\mathrm{et}$ is a free parameter of the label creation and not the acquisition slice thickness: it controls how far the 2D contour information is extruded along the plane normal and thereby the number of labeled voxels per annotation. Larger values of $\mathrm{et}$ yield denser labels at the cost of a stronger assumption of local vessel-shape constancy along the normal direction; the resulting trade-off is quantified in \ref{app:coord}.

\subsubsection{Joining the Intermediate Label Maps}
For each time frame $\tau$, the set of intermediate label maps is
\begin{equation}
    \mathcal{Y}^{(\tau)} = \bigl\{ Y_{\mathcal{S}} \bigr\} \cup
    \bigl\{ Y_{\Pi_m}^{(\tau)} \mid m = 1, \ldots, M \bigr\},
\end{equation}
and they are joined by a consistency rule: a voxel is assigned class $c$ only if every intermediate label map assigns either $c$ or $\xi$; in all other cases (contradicting classes, or $\xi$ in all maps) the voxel is set to $\xi$:
\begin{equation}
Y^{(\tau)}(\mathbf{v}) =
\begin{cases}
c, & \text{if } \exists\, c \in C \mid \forall\, Y_i \in \mathcal{Y}^{(\tau)},\;
      Y_i(\mathbf{v}) \in \{c, \xi\}, \\
\xi, & \text{else}.
\end{cases}
\end{equation}
Conflicts may occur where the extruded regions of two neighboring annotation planes overlap and their contours disagree slightly at the luminal boundary, or where a centerline label contradicts a contour label. Assigning $\xi$ in these cases guarantees that no erroneous supervision signal is introduced.

\subsection{Concatenation to the 4D Label Map}
The final sparse 4D label map is obtained by concatenating the per-frame label maps along the temporal axis:
\begin{equation}
    Y : \mathbb{D}_{4} \rightarrow C \cup \{\xi\}, \qquad
    Y(\mathbf{v}, \tau) = Y^{(\tau)}(\mathbf{v}).
\end{equation}
Voxels that are labeled by the centerline rule are constant over time, whereas voxels that are labeled by the cross-sectional rule vary over time and thus provide the temporal supervision signal for the 4D network. No temporal smoothing or interpolation between frames is applied, i.e.\ every annotated time frame contributes independently.

\section{Coordinate-Descent Selection of Model and Sparse-Label Configuration}
\label{app:coord}

Table~\ref{tab:supp_coordinate_descent} lists all configurations evaluated in the four coordinate-descent steps summarized in Section~\ref{sec:selection}. All values were obtained on the subject-wise 5-fold cross-validation of the training set.

\begin{table}[H]
\centering
\caption{All four steps of the coordinate-descent selection of model configuration and sparse-label parameters. All values were obtained on the subject-wise 5-fold cross-validation of the internal training set and are reported as mean\,$\pm$\,standard deviation over all \num{90354} annotated 2D contours (\num{2954} cross-sections $\times$ the number of reconstructed time frames of the respective scan). The configuration selected in each step is in \textbf{bold}. Step~1 was performed with $\text{et}=\SI{4}{\milli\meter}$ and $\text{cr}=\SI{4}{\milli\meter}$, step~2 with the hypercube kernel, 32 channels and temporal resampling, step~3 with $\text{et}=\SI{6}{\milli\meter}$ and $\text{cr}=\SI{4}{\milli\meter}$, and step~4 with the hybrid kernel, 32 channels and temporal resampling. In step~4 the label configuration selected in step~2 was re-selected, i.e.\ the coordinate descent converged.}
\label{tab:supp_coordinate_descent}
\scriptsize
\setlength{\tabcolsep}{4pt}
\begin{threeparttable}
\sisetup{
  table-number-alignment = center,
  separate-uncertainty   = true,
  detect-weight          = true,
  detect-inline-weight   = math
}
\begin{tabular}{@{}
    c l c
    S[table-format=2.0]
    S[table-format=1.4(3)]
    S[table-format=1.3(3)]
    S[table-format=1.3(3)]
    S[table-format=1.3]@{}}
\toprule
\textbf{Step} & \textbf{Search axis} & \textbf{Temp.\ res.}
& {\textbf{Chan.}} & {\textbf{DSC}} & {\textbf{HD}} & {\textbf{ACD}}
& {\textbf{Failed}} \\
& \textbf{configuration} & & & & {\si{\milli\meter}} & {\si{\milli\meter}} & {\si{\percent}} \\
\midrule
\multirow{5}{*}{1}
 & $3\!\times\!3\!\times\!3$ (3D)                    & No  & 32 & 0.9136 \pm 0.068 & 2.702 \pm 1.395 & 0.826 \pm 0.601 & 0.270 \\
 & $3\!+\!3\!\times\!3\!\times\!3$ (hybrid)          & Yes & 32 & 0.9230 \pm 0.063 & 2.461 \pm 1.158 & 0.727 \pm 0.500 & 0.282 \\
 & $3\!+\!3\!\times\!3\!\times\!3$ (hybrid)          & No  & 32 & 0.9229 \pm 0.060 & 2.472 \pm 1.140 & 0.730 \pm 0.510 & 0.240 \\
 & $3\!\times\!3\!\times\!3\!\times\!3$ (hypercube)  & Yes & 10 & 0.9228 \pm 0.059 & 2.465 \pm 1.120 & 0.729 \pm 0.503 & 0.235 \\
 & \textbf{$\boldsymbol{3\!\times\!3\!\times\!3\!\times\!3}$ (hypercube)} & \textbf{Yes} & \bfseries 32 & \bfseries 0.9236 \pm 0.058 & \bfseries 2.444 \pm 1.165 & \bfseries 0.731 \pm 0.509 & \bfseries 0.207 \\
\midrule
\multirow{6}{*}{2}
 & et = \SI{2}{\milli\meter}, cr = \SI{2}{\milli\meter} & Yes & 32 & 0.9207 \pm 0.062 & 2.474 \pm 1.117 & 0.760 \pm 0.551 & 0.227 \\
 & et = \SI{2}{\milli\meter}, cr = \SI{4}{\milli\meter} & Yes & 32 & 0.9213 \pm 0.051 & 2.530 \pm 1.293 & 0.767 \pm 0.540 & 0.095 \\
 & et = \SI{4}{\milli\meter}, cr = \SI{2}{\milli\meter} & Yes & 32 & 0.9238 \pm 0.059 & 2.419 \pm 1.084 & 0.725 \pm 0.515 & 0.214 \\
 & et = \SI{4}{\milli\meter}, cr = \SI{4}{\milli\meter} & Yes & 32 & 0.9236 \pm 0.058 & 2.444 \pm 1.165 & 0.731 \pm 0.509 & 0.207 \\
 & et = \SI{6}{\milli\meter}, cr = \SI{2}{\milli\meter} & Yes & 32 & 0.9248 \pm 0.057 & 2.408 \pm 1.104 & 0.717 \pm 0.505 & 0.203 \\
 & \textbf{et = \SI{6}{\milli\meter}, cr = \SI{4}{\milli\meter}} & \textbf{Yes} & \bfseries 32 & \bfseries 0.9250 \pm 0.054 & \bfseries 2.423 \pm 1.133 & \bfseries 0.720 \pm 0.507 & \bfseries 0.149 \\
\midrule
\multirow{5}{*}{3}
 & $3\!\times\!3\!\times\!3$ (3D)                    & No  & 32 & 0.9157 \pm 0.064 & 2.658 \pm 1.327 & 0.806 \pm 0.582 & 0.231 \\
 & \textbf{$\boldsymbol{3\!+\!3\!\times\!3\!\times\!3}$ (hybrid)} & \textbf{Yes} & \bfseries 32 & \bfseries 0.9251 \pm 0.054 & \bfseries 2.435 \pm 1.205 & \bfseries 0.718 \pm 0.519 & \bfseries 0.154 \\
 & $3\!+\!3\!\times\!3\!\times\!3$ (hybrid)          & No  & 32 & 0.9236 \pm 0.061 & 2.449 \pm 1.114 & 0.722 \pm 0.499 & 0.259 \\
 & $3\!\times\!3\!\times\!3\!\times\!3$ (hypercube)  & Yes & 10 & 0.9236 \pm 0.060 & 2.440 \pm 1.110 & 0.720 \pm 0.491 & 0.240 \\
 & $3\!\times\!3\!\times\!3\!\times\!3$ (hypercube)  & Yes & 32 & 0.9250 \pm 0.054 & 2.423 \pm 1.133 & 0.720 \pm 0.507 & 0.149 \\
\midrule
\multirow{6}{*}{4}
 & et = \SI{2}{\milli\meter}, cr = \SI{2}{\milli\meter} & Yes & 32 & 0.9202 \pm 0.067 & 2.476 \pm 1.120 & 0.755 \pm 0.540 & 0.311 \\
 & et = \SI{2}{\milli\meter}, cr = \SI{4}{\milli\meter} & Yes & 32 & 0.9201 \pm 0.063 & 2.542 \pm 1.229 & 0.762 \pm 0.532 & 0.270 \\
 & et = \SI{4}{\milli\meter}, cr = \SI{2}{\milli\meter} & Yes & 32 & 0.9231 \pm 0.065 & 2.426 \pm 1.060 & 0.724 \pm 0.496 & 0.312 \\
 & et = \SI{4}{\milli\meter}, cr = \SI{4}{\milli\meter} & Yes & 32 & 0.9230 \pm 0.063 & 2.461 \pm 1.158 & 0.727 \pm 0.500 & 0.282 \\
 & et = \SI{6}{\milli\meter}, cr = \SI{2}{\milli\meter} & Yes & 32 & 0.9239 \pm 0.063 & 2.413 \pm 1.099 & 0.717 \pm 0.492 & 0.292 \\
 & \textbf{et = \SI{6}{\milli\meter}, cr = \SI{4}{\milli\meter}} & \textbf{Yes} & \bfseries 32 & \bfseries 0.9251 \pm 0.054 & \bfseries 2.435 \pm 1.205 & \bfseries 0.718 \pm 0.519 & \bfseries 0.154 \\
\bottomrule
\end{tabular}
\begin{tablenotes}[flushleft]\scriptsize
\item et: extrusion thickness; cr: centerline radius; Temp.\ res.: temporal resampling to 32 frames; Chan.: number of feature maps in the first layer; DSC: Dice similarity coefficient; HD: Hausdorff distance; ACD: average contour distance. In each step the configuration with the highest mean DSC was selected.
\end{tablenotes}
\end{threeparttable}
\end{table}

\section{Agreement Between 2D+time and 4D Volumetric Evaluation}
\label{app:2dt_vs_4d}

\begin{table}[H]
\centering
\scriptsize
\setlength{\tabcolsep}{4pt}
\caption{Agreement between case-wise 2D+time cross-sectional and 4D volumetric evaluation metrics, assessed on $n=36$ automatic segmentations (6 methods $\times$ 6 cases). Bias and limits of agreement (LoA) are reported as $\text{metric}_{\text{2Dt}}-\text{metric}_{\text{4D}}$.}
\label{tab:4D_segmentation_agreement}
\begin{tabular}{lccccc}
\toprule
Metric pair & Bias & 95\,\% LoA & ICC(2,1) [95\,\% CI] & Pearson $r$ & Spearman $\rho$ \\
\midrule
$\text{mean}(\text{DSC}_{\text{2Dt}})$ vs.\ $\text{DSC}_{\text{4D}}$
 & $0.012$ & $[-0.048,0.072]$ & $0.985$ [$0.97$, $0.99$] & $0.986$ & $0.848$ \\
$\text{mean}(\text{ACD}_{\text{2Dt}})$ vs.\ $\text{ASD}_{\text{4D}}$
 & $-0.153$ & $[-0.869,0.562]$ & $0.741$ [$0.53$, $0.86$] & $0.785$ & $0.803$ \\
$\max(\text{HD}_{\text{2Dt}})$ vs.\ $\text{HD95}_{\text{4D}}$
 & $-7.06$ & $[-62.27,48.15]$ & $0.156$ [$-0.16$, $0.45$] & $0.398$ & $0.780$ \\
\bottomrule
\end{tabular}
\end{table}

Table~\ref{tab:4D_segmentation_agreement} reports the agreement between the case-wise 2D+time and 4D evaluation metrics on the six 4D-annotated internal test cases evaluated with six segmentation methods. Biases are the averaged paired differences between the 2D+time and the 4D metric. As the 2D+time and the 4D reference annotations were created by different experts using different annotation techniques (Section~\ref{sec:annotation}), the reported agreement is therefore additionally limited by inter-observer variability and cannot reach unity. DSC showed excellent absolute agreement (\(\mathrm{ICC}=0.985\)), with a mean bias of \(+0.012\), i.e.\ the 2D+time DSC was slightly higher than the corresponding 4D DSC. The 2D+time average contour distance and the 4D average surface distance showed moderate agreement (\(\mathrm{ICC}=0.741\); bias \(-0.15\)~\si{\milli\meter}), whereas \(\max(\mathrm{HD}_{\mathrm{2Dt}})\) and \(\mathrm{HD95}_{\mathrm{4D}}\) showed poor absolute agreement (\(\mathrm{ICC}=0.156\); bias \(-7.06\)~\si{\milli\meter}) with a monotonic association (\(\rho=0.780\)). This is reasonable, as cross-sections without a predicted lumen are excluded from \(\max(\mathrm{HD}_{\mathrm{2Dt}})\), whereas the corresponding missing volume can cause a large \(\mathrm{HD95}_{\mathrm{4D}}\). Within this limited subset, and within the bounds set by inter-observer variability, the results support interpreting the 2D+time DSC and the average-distance measure as proxies for their 4D counterparts, while the 2D+time HD should be interpreted as a secondary relative-error measure only.

\begin{figure}[H]
    \centering
    \includegraphics[width=1\linewidth]{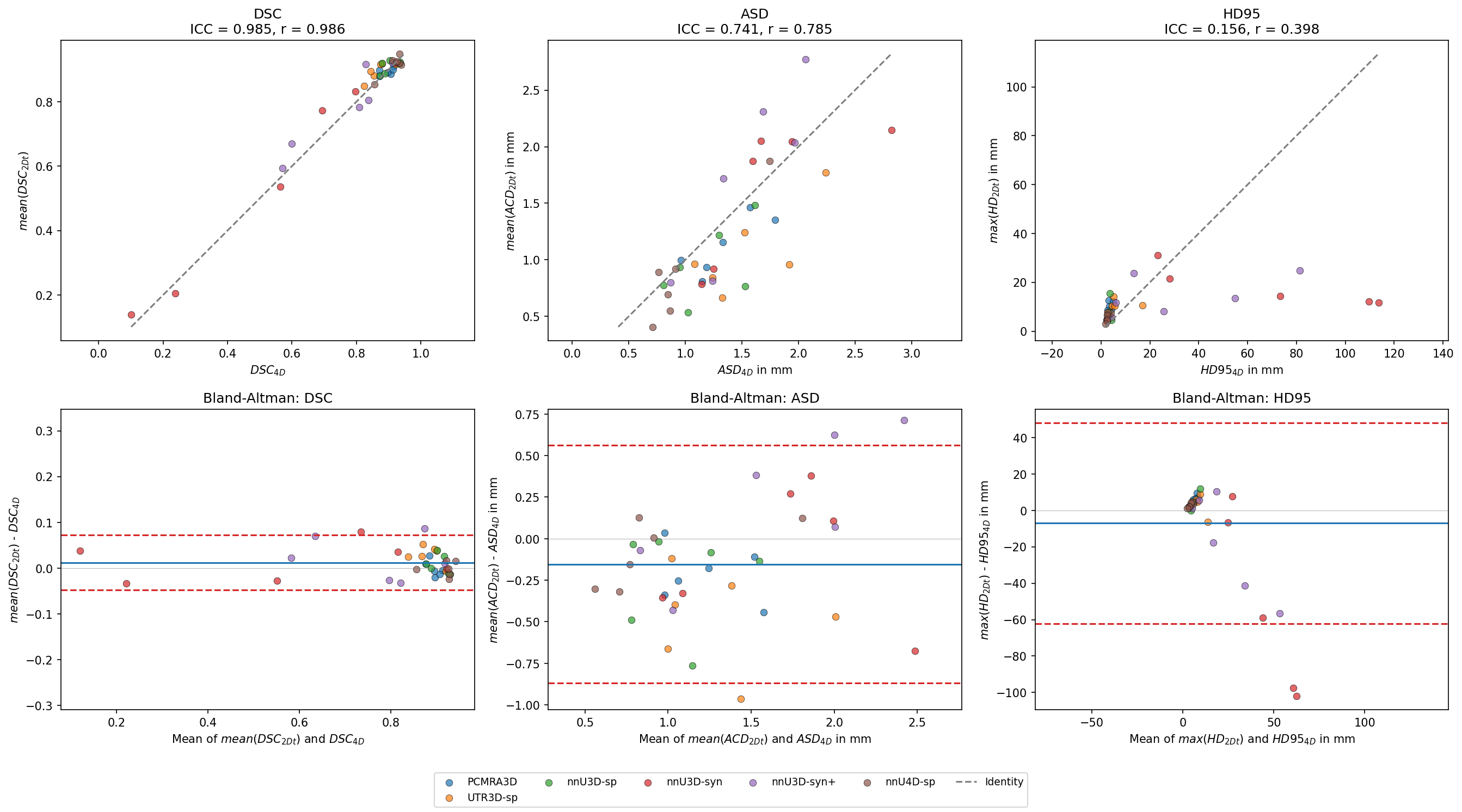}
    \caption{Agreement between case-wise 2D+time cross-sectional and 4D volumetric evaluation metrics, assessed on $n=36$ automatic segmentations (6 methods $\times$ 6 cases of the internal test set for which a 4D expert annotation was available). Colors indicate the segmentation method (see legend). The six cases were deliberately selected to span the full range of segmentation quality and are therefore not representative.}
    \label{fig:2dt_vs_4d_agreement}
\end{figure}

Figure~\ref{fig:2dt_vs_4d_agreement} shows the scatter and Bland-Altman plots underlying the summary statistics of Table~\ref{tab:4D_segmentation_agreement}.

\section{Segmentation Performance per Cross-Sectional Position}
\label{app:perposition}

Table~\ref{tab:supp_cross_section_performance} and Table~\ref{tab:supp_cross_section_performance_ets} report all segmentation metrics separately for each annotated cross-sectional position, so that method behavior can be compared at positions at which all methods succeed. In line with the evaluation protocol, contours without a predicted lumen were counted as failures and entered the DSC with $\text{DSC}=0$, but were excluded from HD and ACD. Distance metrics of methods with a high failure rate are therefore computed on an easier subset of contours and must not be compared in isolation.

\begin{landscape}
\begin{table}[p]
\centering
\caption{Segmentation performance per cross-sectional position for all evaluated methods on the test set. Reported are the mean Dice similarity coefficient (DSC), mean Hausdorff distance (HD), mean average contour distance (ACD), and the number of failed contours out of the $n$ evaluated 2D contours at that position ($n$ = number of cross-sections at that position $\times$ the number of reconstructed time frames of the respective scan).}
\label{tab:supp_cross_section_performance}
\scriptsize
\setlength{\tabcolsep}{3pt}
\begin{adjustbox}{max width=\linewidth}
\begin{tabular}{ll
cccc
cccc
cccc
cccc
cccc
cccc}
\toprule
\multirow{2}{*}{Cross-section} & \multirow{2}{*}{$n$}
& \multicolumn{4}{c}{PCMRA3D}
& \multicolumn{4}{c}{nnU3D-syn}
& \multicolumn{4}{c}{nnU3D-syn+}
& \multicolumn{4}{c}{UTR3D-sp}
& \multicolumn{4}{c}{nnU3D-sp}
& \multicolumn{4}{c}{nnU4D-sp} \\
\cmidrule(lr){3-6}
\cmidrule(lr){7-10}
\cmidrule(lr){11-14}
\cmidrule(lr){15-18}
\cmidrule(lr){19-22}
\cmidrule(lr){23-26}
&
& DSC & HD & ACD & Failed
& DSC & HD & ACD & Failed
& DSC & HD & ACD & Failed
& DSC & HD & ACD & Failed
& DSC & HD & ACD & Failed
& DSC & HD & ACD & Failed \\
\midrule
A3.1 & 923
& 0.866 & 5.772 & 1.613 & 0
& 0.229 & 8.399 & 3.107 & 647
& 0.648 & 6.659 & 2.596 & 172
& 0.870 & 4.718 & 1.496 & 3
& 0.895 & 3.935 & 1.219 & 0
& 0.913 & 3.501 & 1.011 & 0 \\
A3.2 & 918
& 0.899 & 4.621 & 1.282 & 0
& 0.362 & 6.945 & 2.847 & 494
& 0.742 & 6.525 & 2.545 & 89
& 0.875 & 3.876 & 1.212 & 31
& 0.914 & 3.474 & 1.078 & 0
& 0.929 & 3.077 & 0.892 & 0 \\
A3.3 & 943
& 0.903 & 3.938 & 1.161 & 0
& 0.459 & 5.354 & 2.075 & 422
& 0.765 & 5.439 & 2.125 & 82
& 0.887 & 3.166 & 0.939 & 33
& 0.927 & 2.852 & 0.815 & 0
& 0.936 & 2.570 & 0.718 & 0 \\
B1 & 943
& 0.891 & 5.191 & 1.361 & 0
& 0.539 & 4.265 & 1.605 & 353
& 0.782 & 4.108 & 1.418 & 106
& 0.880 & 3.338 & 0.996 & 33
& 0.926 & 2.717 & 0.809 & 0
& 0.931 & 2.623 & 0.788 & 0 \\
B2 & 783
& 0.867 & 5.206 & 1.415 & 0
& 0.559 & 4.411 & 1.512 & 275
& 0.774 & 4.081 & 1.417 & 87
& 0.876 & 2.919 & 0.828 & 33
& 0.926 & 2.599 & 0.723 & 0
& 0.927 & 2.602 & 0.731 & 0 \\
B3 & 811
& 0.870 & 3.824 & 1.236 & 0
& 0.660 & 4.075 & 1.665 & 183
& 0.736 & 3.819 & 1.621 & 114
& 0.873 & 2.651 & 0.811 & 33
& 0.918 & 2.458 & 0.757 & 0
& 0.927 & 2.179 & 0.667 & 0 \\
B4.1 & 943
& 0.881 & 3.947 & 1.094 & 0
& 0.663 & 3.900 & 1.714 & 204
& 0.716 & 3.808 & 1.630 & 148
& 0.877 & 2.582 & 0.788 & 33
& 0.915 & 2.436 & 0.758 & 0
& 0.928 & 2.206 & 0.634 & 0 \\
B4.2 & 914
& 0.893 & 3.567 & 0.931 & 0
& 0.714 & 3.917 & 1.722 & 138
& 0.751 & 3.635 & 1.568 & 108
& 0.873 & 2.568 & 0.865 & 33
& 0.916 & 2.515 & 0.771 & 0
& 0.923 & 2.231 & 0.680 & 0 \\
B4.3 & 918
& 0.908 & 3.015 & 0.810 & 0
& 0.725 & 3.604 & 1.364 & 148
& 0.784 & 3.272 & 1.257 & 94
& 0.883 & 2.447 & 0.756 & 33
& 0.925 & 2.162 & 0.632 & 0
& 0.929 & 2.095 & 0.639 & 0 \\
D1.1 & 918
& 0.912 & 3.132 & 0.839 & 0
& 0.705 & 3.719 & 1.259 & 179
& 0.805 & 3.313 & 1.237 & 77
& 0.909 & 2.708 & 0.803 & 0
& 0.923 & 2.235 & 0.675 & 0
& 0.929 & 2.097 & 0.642 & 0 \\
D1.2 & 918
& 0.908 & 2.983 & 0.831 & 0
& 0.540 & 3.836 & 1.449 & 337
& 0.751 & 3.408 & 1.364 & 117
& 0.906 & 2.690 & 0.782 & 0
& 0.920 & 2.298 & 0.666 & 0
& 0.923 & 2.213 & 0.654 & 0 \\
D1.3 & 893
& 0.917 & 2.638 & 0.683 & 0
& 0.486 & 3.643 & 1.314 & 389
& 0.794 & 3.345 & 1.243 & 76
& 0.908 & 2.574 & 0.795 & 0
& 0.921 & 2.240 & 0.635 & 0
& 0.926 & 2.095 & 0.610 & 0 \\
\bottomrule
\end{tabular}
\end{adjustbox}
\end{table}
\end{landscape}

\begin{landscape}
\begin{table}[!p]
\centering
\caption{Segmentation performance per cross-sectional position for all evaluated methods on the external test set. Reported metrics are the mean Dice similarity coefficient (DSC), mean Hausdorff distance (HD), mean average contour distance (ACD), and the number of failed segmentations (Failed). The number of evaluated 2D annotations per cross-sectional position ($n$) is shown separately. Cross-sections in the ascending aorta were defined at the aortic annulus (Annulus) and at \SI{20}{\milli\meter} increments distal to the annulus (Annulus +20 mm to Annulus +160 mm). Descending aortic cross-sections were defined at the level of the pulmonary artery (Desc.\ aorta (PA level)) and in the distal descending aorta (Distal desc.\ aorta).}
\label{tab:supp_cross_section_performance_ets}
\scriptsize
\setlength{\tabcolsep}{3pt}
\begin{adjustbox}{max width=\linewidth}
\begin{tabular}{llcccccccccccccccccccccccc}
\toprule
Cross-section & $n$
& \multicolumn{4}{c}{ManReg4D}
& \multicolumn{4}{c}{nnU3D-syn}
& \multicolumn{4}{c}{nnU3D-syn+}
& \multicolumn{4}{c}{UTR3D-sp}
& \multicolumn{4}{c}{nnU3D-sp}
& \multicolumn{4}{c}{nnU4D-sp} \\
\cmidrule(lr){3-6}
\cmidrule(lr){7-10}
\cmidrule(lr){11-14}
\cmidrule(lr){15-18}
\cmidrule(lr){19-22}
\cmidrule(lr){23-26}
&
& DSC & HD & ACD & Failed
& DSC & HD & ACD & Failed
& DSC & HD & ACD & Failed
& DSC & HD & ACD & Failed
& DSC & HD & ACD & Failed
& DSC & HD & ACD & Failed \\
\midrule
Annulus & 805 & 0.840 & 5.509 & 1.969 & 0 & 0.279 & 11.175 & 4.397 & 492 & 0.731 & 7.054 & 2.474 & 93 & 0.610 & 6.444 & 2.359 & 201 & 0.832 & 3.997 & 1.295 & 54 & 0.900 & 3.394 & 0.946 & 12 \\
Annulus +20 mm & 805 & 0.852 & 5.078 & 1.773 & 0 & 0.397 & 9.254 & 3.312 & 391 & 0.772 & 6.984 & 2.512 & 49 & 0.677 & 5.249 & 1.844 & 159 & 0.868 & 3.541 & 1.181 & 26 & 0.936 & 2.647 & 0.705 & 0 \\
Annulus +40 mm & 805 & 0.847 & 5.457 & 1.801 & 0 & 0.442 & 7.434 & 2.728 & 363 & 0.758 & 6.112 & 2.265 & 82 & 0.739 & 4.266 & 1.370 & 128 & 0.894 & 3.314 & 0.989 & 10 & 0.933 & 2.637 & 0.689 & 0 \\
Annulus +60 mm & 780 & 0.836 & 4.978 & 1.761 & 0 & 0.525 & 5.804 & 2.157 & 280 & 0.759 & 5.579 & 2.332 & 65 & 0.726 & 4.409 & 1.431 & 126 & 0.865 & 3.430 & 1.017 & 29 & 0.931 & 2.466 & 0.675 & 0 \\
Annulus +80 mm & 725 & 0.804 & 5.520 & 1.924 & 0 & 0.606 & 7.057 & 2.639 & 165 & 0.744 & 5.762 & 2.741 & 41 & 0.776 & 4.004 & 1.290 & 72 & 0.870 & 3.247 & 1.059 & 14 & 0.915 & 2.713 & 0.762 & 0 \\
Annulus +100 mm & 420 & 0.792 & 5.370 & 2.136 & 0 & 0.686 & 6.138 & 2.692 & 55 & 0.754 & 5.676 & 2.741 & 19 & 0.866 & 3.693 & 1.174 & 3 & 0.900 & 3.045 & 0.935 & 0 & 0.922 & 2.438 & 0.695 & 0 \\
Annulus +120 mm & 170 & 0.786 & 5.584 & 2.219 & 0 & 0.758 & 5.169 & 1.854 & 17 & 0.818 & 5.457 & 2.294 & 0 & 0.858 & 3.696 & 1.192 & 3 & 0.901 & 3.000 & 0.901 & 0 & 0.920 & 2.670 & 0.743 & 0 \\
Annulus +140 mm & 25 & 0.872 & 3.786 & 1.271 & 0 & 0.891 & 3.769 & 1.368 & 0 & 0.859 & 4.997 & 1.522 & 0 & 0.912 & 3.062 & 0.789 & 0 & 0.904 & 2.897 & 0.668 & 0 & 0.909 & 2.663 & 0.608 & 0 \\
Annulus +160 mm & 25 & 0.881 & 3.695 & 1.129 & 0 & 0.910 & 3.239 & 1.089 & 0 & 0.851 & 4.927 & 1.522 & 0 & 0.925 & 2.638 & 0.705 & 0 & 0.927 & 2.324 & 0.661 & 0 & 0.939 & 2.113 & 0.545 & 0 \\
Desc.\ aorta (PA level) & 805 & 0.800 & 4.292 & 1.688 & 0 & 0.678 & 4.161 & 1.645 & 155 & 0.675 & 6.513 & 2.434 & 125 & 0.725 & 3.862 & 1.534 & 97 & 0.823 & 2.576 & 0.846 & 69 & 0.876 & 2.404 & 0.872 & 10 \\
Distal desc.\ aorta & 805 & 0.685 & 4.421 & 2.049 & 30 & 0.399 & 4.090 & 1.569 & 416 & 0.295 & 4.106 & 1.697 & 517 & 0.376 & 4.847 & 2.300 & 381 & 0.741 & 4.323 & 1.720 & 70 & 0.879 & 2.254 & 0.695 & 23 \\
\bottomrule
\end{tabular}
\end{adjustbox}
\end{table}
\end{landscape}

\section{Phase-Resolved Segmentation Performance by Dataset Characteristics}
\label{app:phase}

\begin{figure}[h]
    \centering
    \includegraphics[width=0.85\linewidth]{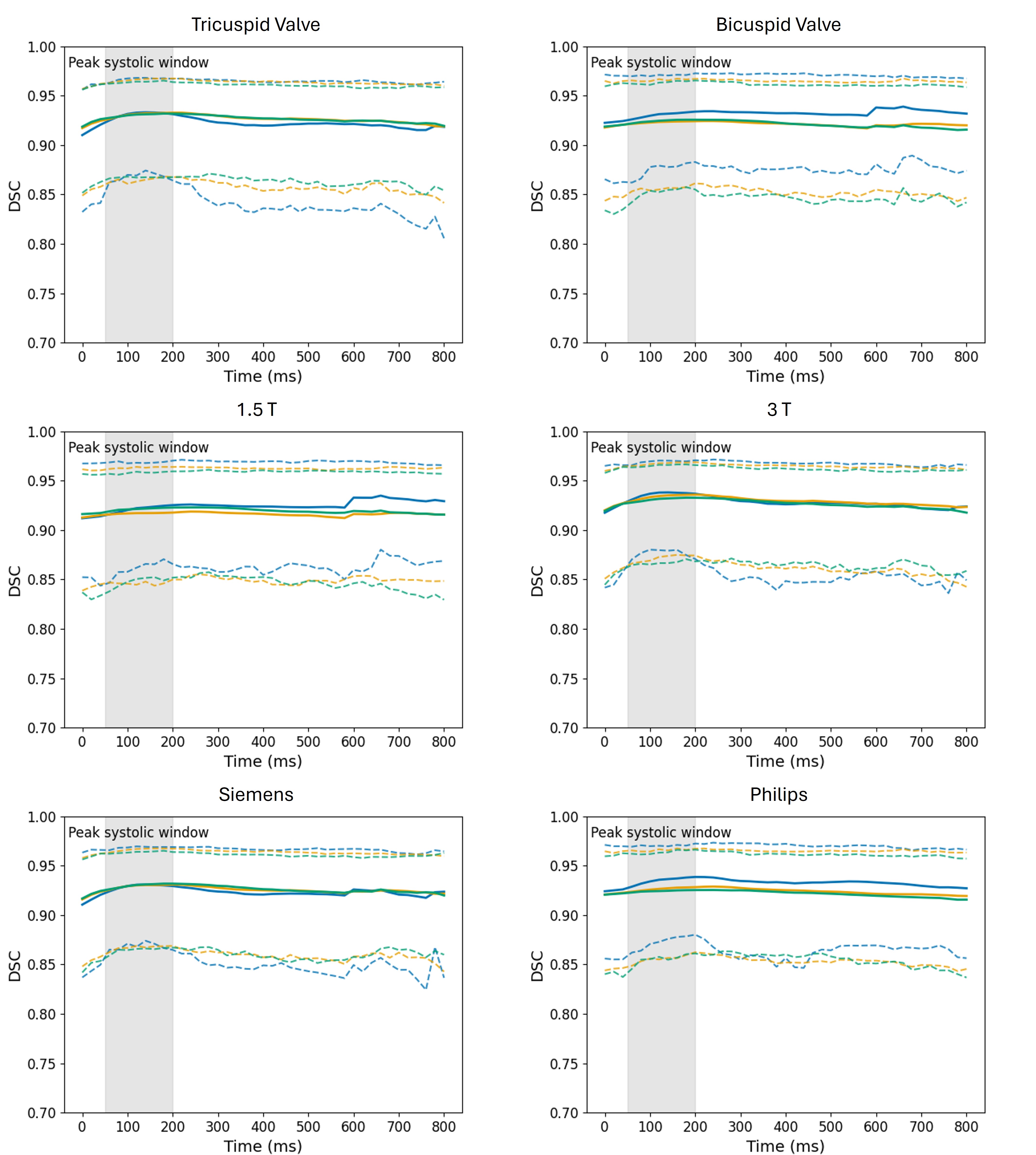}
    \caption{Dice similarity coefficient (DSC) of the proposed 4D U-Net over the cardiac cycle, stratified by dataset characteristics. The solid line shows the mean DSC and the dashed lines the 5th and 95th percentiles.}
    \label{fig:dice_over_time_characteristics}
\end{figure}

\section{Bland-Altman Analyses of the Quantitative Parameters}
\label{app:bland_altman}

\begin{figure}
    \centering
    \includegraphics[width=0.85\linewidth]{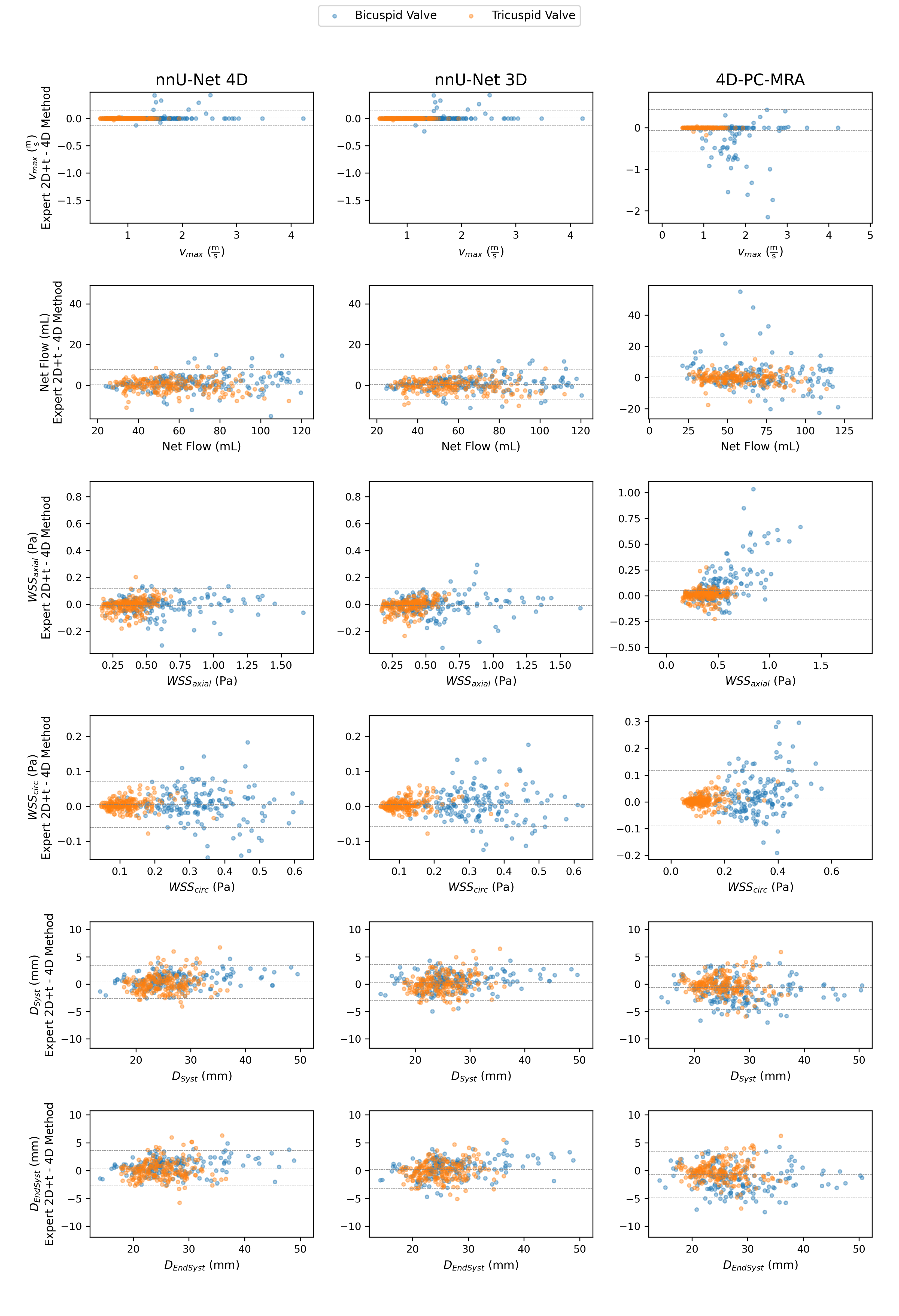}
    \caption{Bland-Altman analysis of the quantitative parameters on the test set. Colors indicate valve configuration (see legend). The solid and dashed lines mark the bias and the $\pm 1.96$ standard deviations of the differences of the proposed nnU4D-sp. Each point corresponds to one annotated 2D+time cross-section. Only the three methods with the lowest failure rate of the respective test set are shown, and all values are computed on the cross-sections that were successfully segmented by all shown methods.}
    \label{fig:bland_altman_test_set}
\end{figure}

\begin{figure}
    \centering
    \includegraphics[width=0.95\linewidth]{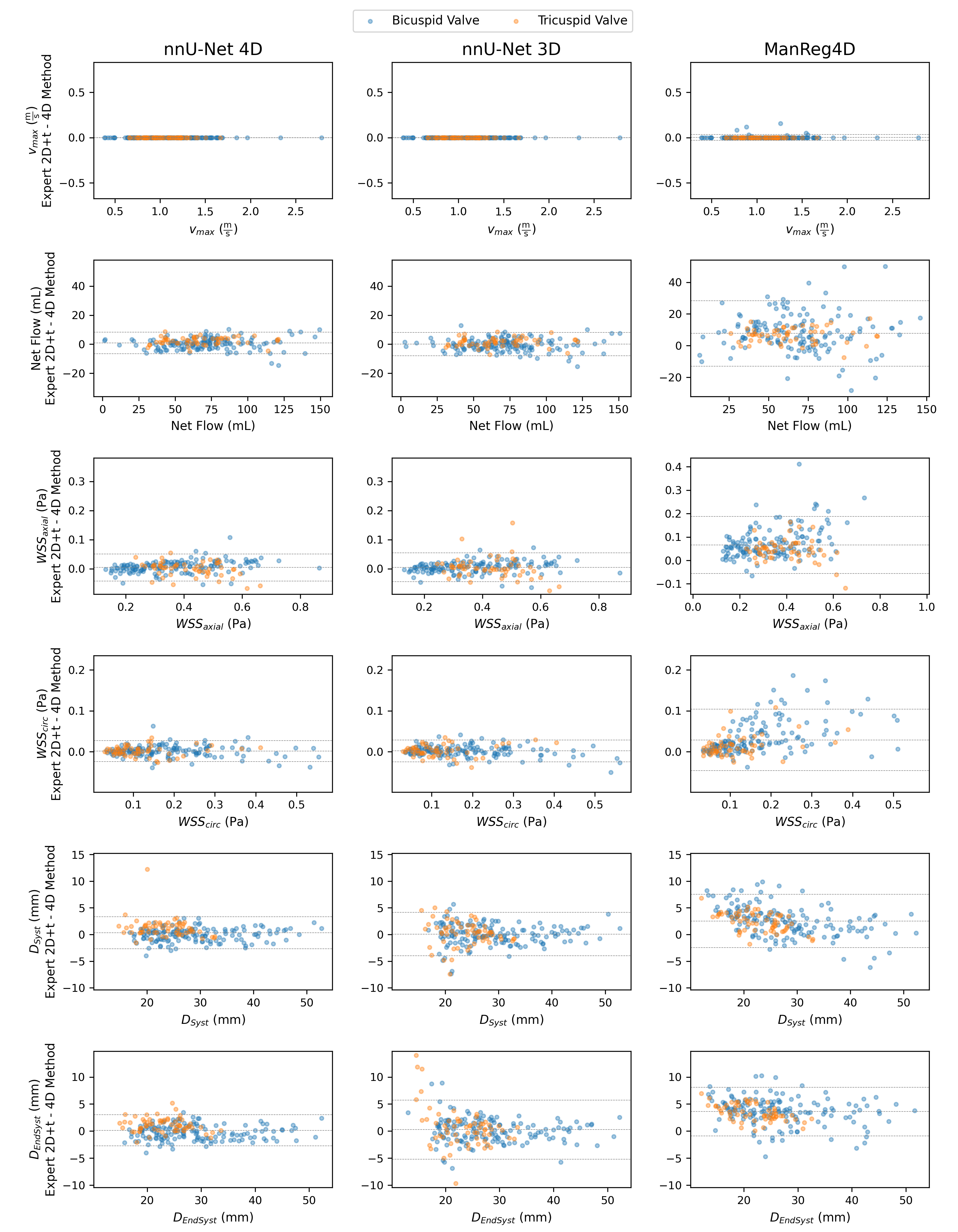}
    \caption{Bland-Altman analysis of the quantitative parameters on the external test set. Plot layout, parameters, and inclusion criteria are identical to Figure~\ref{fig:bland_altman_test_set}. Note that $v_{\max}$ is identical for the compared methods in most cross-sections, because the location of the peak through-plane velocity lies well inside the lumen and is therefore insensitive to the boundary differences between the segmentations.}
    \label{fig:bland_altman_external_test_set}
\end{figure}

Figure~\ref{fig:bland_altman_test_set} and Figure~\ref{fig:bland_altman_external_test_set} show the Bland-Altman analyses underlying the summary statistics of Table~\ref{tab:quantitative_parameter_evaluation} of the main text.

\end{document}